\documentclass[10pt,twocolumn,letterpaper]{article}

\usepackage[pagenumbers]{cvpr} 

\usepackage{graphicx}
\usepackage{amsmath}
\usepackage{amssymb}
\usepackage{pifont}
\usepackage{booktabs}
\usepackage{xcolor}
\usepackage{multirow,tabularx}
\usepackage{makecell}

\usepackage[accsupp]{axessibility}  

\usepackage[pagebackref,breaklinks,colorlinks]{hyperref}

\usepackage[capitalize]{cleveref}
\crefname{section}{Sec.}{Secs.}
\Crefname{section}{Section}{Sections}
\Crefname{table}{Table}{Tables}
\crefname{table}{Tab.}{Tabs.}

\newcommand{\dataname}{\textsc{HiREST}}

\newcommand{\videoretrievaltask}{video retrieval}
\newcommand{\momentretrievaltask}{moment retrieval}
\newcommand{\stepretrievalsubtask}{moment segmentation}
\newcommand{\momentcaptioningtask}{step captioning}

\newcommand{\Videoretrievaltask}{Video retrieval}
\newcommand{\Momentretrievaltask}{Moment retrieval}
\newcommand{\Stepretrievalsubtask}{Moment segmentation}
\newcommand{\Momentcaptioningtask}{Step captioning}

\newcommand{\cmark}{\ding{51}}

\def\cvprPaperID{12156} 
\def\confName{CVPR}
\def\confYear{2023}

\begin{document}

\title{
Hierarchical Video-Moment Retrieval and Step-Captioning
}

\newcommand*\samethanks[1][\value{footnote}]{\footnotemark[#1]}

\author{
  Abhay Zala\thanks{equal contribution}\ \ $^1$ \qquad
  Jaemin Cho\samethanks{}\ \ $^1$ \qquad
  Satwik Kottur$^2$ \qquad
  Xilun Chen$^2$  \\
  Barlas Oguz$^2$ \qquad 
  Yashar Mehdad$^2$ \qquad
  Mohit Bansal$^1$ \\
  UNC Chapel Hill$^1$ \quad \quad Meta AI$^2$\\
  {\tt\small \{jmincho, aszala, mbansal\}@cs.unc.edu} \quad \quad 
  {\tt\small \{skottur, xilun, barlaso, mehdad\}@fb.com} 
  \\  
  {\tt \normalsize \href{https://hirest-cvpr2023.github.io}{https://hirest-cvpr2023.github.io}}
}
\maketitle

\begin{abstract}
There is growing interest in searching for information from large video corpora.
Prior works have studied relevant tasks, such as text-based video retrieval, moment retrieval, video summarization, and video captioning in isolation,
without an end-to-end setup that can jointly search from video corpora and generate summaries.
Such an end-to-end setup would allow for many interesting applications,
\eg, a text-based search that finds a relevant video from a video corpus, extracts the most relevant moment from that video, and segments the moment into important steps with captions.
To address this,
we present the \dataname{}
(\textbf{HI}erarchical \textbf{RE}trieval and \textbf{ST}ep-captioning) dataset
and propose a new benchmark that covers hierarchical information retrieval and visual/textual stepwise summarization from an instructional video corpus.
\dataname{} consists of 3.4K
text-video pairs from an instructional video dataset, 
where 1.1K videos have annotations of moment spans relevant to text query and breakdown of each moment into key instruction steps with caption and timestamps (totaling 8.6K step captions).
Our hierarchical benchmark consists of
video retrieval, moment retrieval, and two novel
moment segmentation and step captioning tasks.
In moment segmentation, models break down a video moment
into instruction steps and identify start-end boundaries.
In step captioning, models generate a textual summary
for each step.
We also present starting point task-specific and end-to-end joint baseline models for our new benchmark.
While the baseline models show some promising results, there still exists large room for future improvement by the community.
\end{abstract}

\section{Introduction}
\label{sec:intro}

\begin{figure*}[t]
    \centering
    \includegraphics[width=.99\textwidth]{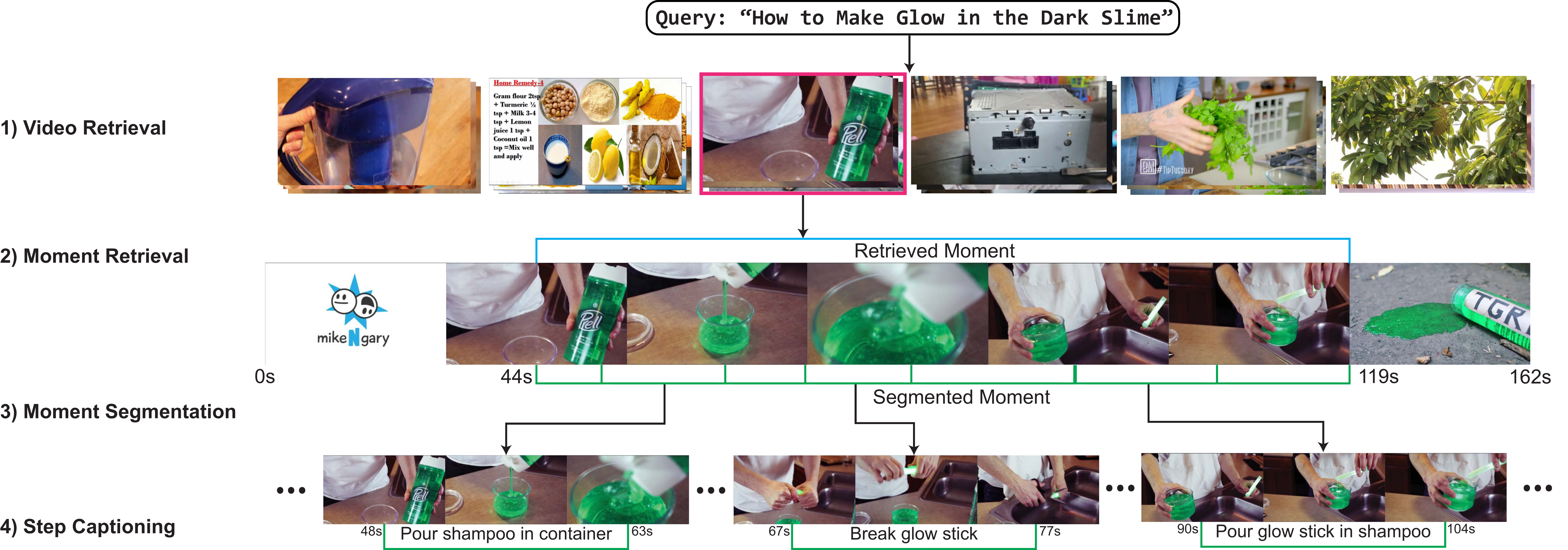}
    \caption{
        Overview of four hierarchical tasks of our \dataname{} dataset (\cref{sec:dataset}).
        1) \Videoretrievaltask{}: find a video that is most relevant to a given text query.
        2) \Momentretrievaltask{}: choose the relevant span of the video, by trimming the parts irrelevant to the text query.
        3) \Stepretrievalsubtask{}: break down the span into several steps and identify the start-end boundaries of each step.
        4) \Momentcaptioningtask{}: generate step-by-step textual summaries of the moment.
    }
    \label{fig:mainfigure}
\end{figure*}

Encouraged by the easy access to smartphones, recording software, and video hosting platforms, people are increasingly accumulating videos of all kinds.
To fuel the subsequent growing interest in using machine learning systems to extract and summarize important information 
from these large video corpora based on text queries,
progress has been made in
video retrieval~\cite{Xu2016MSRVTT,Yu2017,lei2020tvr,Li2020Hero,Bain21FrozenInTime},
moment retrieval~\cite{Hendricks2017LocalizingMI,lei2020tvr,Lei2021QVHighlightsDM},
video summarization~\cite{Gygli2014CreatingSF,Song2015TVSumSW,Sharghi2017QueryFocusedVS,Narasimhan2022TLDWSI}, and
video captioning~\cite{Xu2016MSRVTT,Yu2017,BMT_Iashin_2020,lin2021SwinBERTend-to-end}.
Previous works have generally focused on solving these tasks independently; however, all these tasks share the common goal of retrieving information from a video corpus, at different levels of scales and via different modalities.
Hence, in this work, we introduce a new hierarchical benchmark that combines all four tasks to enable novel and useful real-world applications.
For example,
a text-based search service that finds a relevant video from a large video corpus, extracts the most relevant moment from that video, segments the moment into important steps, and captions them for easy indexing and retrieval.
To support this, we introduce \dataname{}, a hierarchical instructional video dataset for a holistic benchmark of information retrieval from a video corpus 
(see \cref{sec:dataset}).
\dataname{} consists of four annotations:
1) $3.4K$ pairs of text query about open-domain instructions (\eg, \textit{`how to make glow in the dark slime'}) and videos,
2) relevant moment timestamps inside the $1.1K$ videos, where only a part of the video ($<75\%$) is relevant to the text query,
3) moment breakdown in several instructional steps with timestamps ($7.6$ steps per video, total $8.6K$ steps),
and,
4) an manually curated English caption for each step (\eg \textit{`pour shampoo in container'}).
We collect fine-grained step-wise annotations of \dataname{} in a two-step annotation process with online crowdworkers on instructional text-video pairs from the HowTo100M~\cite{miech19howto100m} dataset (see \cref{sec:dataset_collection}).
The instructional videos often come with clear step-by-step instructions, allowing fine-grained segmentation of the videos into short steps.
While there are existing video datasets with step annotations, they are based on a small number of predefined task names~\cite{Zhukov2019CrossTask,Tang2019COIN} (thus step captions are not diverse), or are limited to a single topic (\eg cooking~\cite{Zhou2018TowardsALYouCook2}).
\dataname{} covers various domains and provides diverse step captions with timestamps written by human annotators (see \Cref{tbl:datasetcompare}), presenting new challenging and realistic benchmarks for hierarchical video information retrieval.

Using the \dataname{} dataset, we benchmark four tasks:
1) \videoretrievaltask{},
2) \momentretrievaltask{},
3) \stepretrievalsubtask{},
and
4) \momentcaptioningtask{} (see \cref{fig:mainfigure} and \cref{sec:tasks}).
In the \videoretrievaltask{} task, models have to identify a video that is most relevant to a given text query.
In the \momentretrievaltask{} task, models have to select the relevant span of the video, by trimming the parts irrelevant to the text query (blue boundary in \cref{fig:mainfigure}).
In the \stepretrievalsubtask{} task, models have to break down the relevant portion into several instructional steps and identify the start-end boundaries of each step
(green boundaries in \cref{fig:mainfigure}).
Finally, in the \momentcaptioningtask{} task, models have to generate step captions (\eg \textit{`spray the warm water on carpet'}) of the instructional steps.
To provide good starting points to the community for our new task hierarchy, we show the performance of recent baseline models on \dataname{}. For baselines, we use strong models including
CLIP~\cite{Radford2021CLIP},
EVA-CLIP~\cite{Fang2023EVA},
Frozen-in-Time\cite{Bain21FrozenInTime},
BMT~\cite{BMT_Iashin_2020}, and SwinBERT~\cite{lin2021SwinBERTend-to-end}.
On all four tasks, we find that finetuning models on \dataname{} improve performance; however, there exists a large room to improve performance.
\par
We summarize our contributions in this paper:
1) We present \dataname{} dataset and propose a new benchmark that covers hierarchy in information retrieval and visual/textual summarization from an instructional video corpus.
2) Unlike existing video datasets with step captions based on predefined task names or limited to a single topic, our \dataname{} provides diverse, high-quality step captions with timestamps written by human annotators.
3) We provide a joint baseline model that can perform \momentretrievaltask{},
\stepretrievalsubtask{},
and
\momentcaptioningtask{} with a single architecture.
4) We provide comprehensive dataset analyses and show experiments with baseline models for each task, where there is a large room to improve model performance.
We hope that \dataname{} can foster future work on end-to-end systems for holistic information retrieval and summarization on large video corpus.
In addition,
our manually annotated step captions can also be a good source for training and testing the step-by-step reasoning of large multimodal language models~\cite{Wei2022COT,Zhang2023multimodalCOT}.

\section{Related Work}
\label{sec:related_work}

\subsection{Text-based Information Retrieval from Video}
With growing interest in building machine learning systems to search for useful information from large video corpora via text searches, several lines of work have been proposed.
In text-to-video retrieval, a system finds the most relevant videos from a list of videos with a given text query~\cite{Xu2016MSRVTT,Yu2017,lei2020tvr,Li2020Hero,Bain21FrozenInTime}.
In moment retrieval, a system finds the most relevant moments (usually a few seconds of frame spans), from a single video~\cite{Hendricks2017LocalizingMI,lei2020tvr,Lei2021QVHighlightsDM}.
In query-focused video summarization, which is a text-conditional version of generic video summarization~\cite{Gygli2014CreatingSF,Song2015TVSumSW}, a system finds the most relevant frames from a video with text query~\cite{Sharghi2017QueryFocusedVS,Narasimhan2022TLDWSI}.
In video captioning, a system generates a short textual description of a given video~\cite{Xu2016MSRVTT,Yu2017,BMT_Iashin_2020,lin2021SwinBERTend-to-end}.
While all these tasks share common goals, information retrieval and summarization from a video corpus, previous works have focused on systems that are specialized in a single task.
In this work, we introduce a holistic setup that combines video retrieval, moment retrieval, query-focused video summarization (called moment segmentation), and generating a stepwise textual summary of short clip (called step captioning), so that users can search for the most relevant video, the most relevant moment inside the video, and get the stepwise text summarization of the moment.

\begin{table*}[t]
\centering
\resizebox{.95\textwidth}{!}{
\begin{tabular}{l l l c c c c c}
  \toprule
   Dataset & Domain & Step caption & \makecell{\# Videos / \# Steps} & \makecell{\# Steps \\per Moment} &  \makecell{\# Words \\ per Caption} & \makecell{\# Unique \\ Captions} & \makecell{Avg. Duration (s) \\Video / Step} \\

   \midrule
    COIN~\cite{Tang2019COIN} & Open & Predefined steps & 11.8K  / 46K   & 3.9 & 4.8 & 0.8K & 142 / 14.9  \\
    CrossTask~\cite{Zhukov2019CrossTask} & Open & Predefined steps & 4.7K / 21K  & 7.4 & 2.4 & 0.1K & 297 / 9.6 \\
    YouCook2~\cite{Zhou2018TowardsALYouCook2} & Cooking & Manually written & 2K  / 14K & 7.7 & 8.8 & 13K & 316 / 19.7  \\
    \midrule
    \dataname{} (Ours) & Open & Manually written & 3.4K (1.1K w/ steps) / 8.6K & 7.6 & 4.4 & 7.9K &  263 / 18.9 \\
 \bottomrule
\end{tabular}
}
\caption{
Comparison of \dataname{} and other video datasets with step annotations.
While smaller in terms of the total number of videos than other datasets,
\dataname{} covers various open-domain videos with many step annotations per video and high-quality step captions written by human annotators.
}
\label{tbl:datasetcompare}
\end{table*}

\subsection{Instructional Video Datasets}
Recently, there have also been several efforts towards creating instructional video datasets~\cite{Zhou2018TowardsALYouCook2,wang-etal-2019-youmakeup,Rohrbach2012MP2,Kuehne2014TheLOBreakfast,miech19howto100m,Tang2019COIN,Zhukov2019CrossTask}. While many of these datasets do a good job of providing high-quality instructional videos, they primarily only target a single domain~\cite{Zhou2018TowardsALYouCook2,wang-etal-2019-youmakeup,Rohrbach2012MP2,Kuehne2014TheLOBreakfast}. There have been recent strong efforts towards developing more diverse instructional datasets~\cite{miech19howto100m,Tang2019COIN,Zhukov2019CrossTask}. Datasets like HowTo100M~\cite{miech19howto100m} provide diverse instructional videos but lack specific step-by-step annotations. 
Some previous works such as \cite{Tang2019COIN,Zhukov2019CrossTask} provide step-level annotations for open domain videos, however, are restricted to a set of predefined steps that are reapplied across several videos.
Our \dataname{} dataset provides step annotations on diverse instructional videos, where
all step captions are manually written to answer the input text query by human annotators (see \Cref{tbl:datasetcompare}).

\section{\dataname{}: Hierarchical Retrieval and Step-Captioning Dataset}
\label{sec:dataset}

We present \dataname{}, a video dataset consisting of 3.4K text-video pairs, 1.8K moments, and 8.6K step caption annotations. It covers the hierarchy of video/moment retrieval and stepwise captioning from a diverse instructional video corpus.
Previous step annotations in video datasets used predefined task descriptions with small vocabulary~\cite{Zhukov2019CrossTask,Tang2019COIN} or limited to a single domain (\eg cooking~\cite{Zhou2018TowardsALYouCook2}).
In contrast, the step captions of \dataname{} are manually written by human annotators and cover diverse domains with a large vocabulary (see \Cref{tbl:datasetcompare}).
We describe the data collection process (\cref{sec:dataset_collection}),
dataset analysis (\cref{sec:dataset_analysis}),
and four hierarchical tasks that stem from our dataset (\cref{sec:tasks}).

\subsection{Dataset Collection}
\label{sec:dataset_collection}

In the following, we describe the two-stage data collection process.
In the appendix, we provide screenshots of the data collection interface for each stage and worker qualification process.

\vspace{3pt} \par \noindent\textbf{Stage 1: Video and Moment Retrieval.}
We collect the pairs of text queries and relevant videos from the HowTo100M~\cite{miech19howto100m} dataset.
Since videos were originally automatically collected from YouTube, we ensure that all videos are actually relevant to the query through human annotation.
We employ crowdworkers from Amazon Mechanical Turk\footnote{\url{https://www.mturk.com}} and ask them to label whether or not the video correctly answers/solves the associated text query.

If the video is labeled as relevant to the text query, then we collect relevant `moment' annotation from the video, by asking the crowdworkers to trim the video to the parts that are directly associated with the text (\ie remove video parts unrelated to the text query, such as intro or other topics).
We define a video as \textit{clippable} to a moment, if the moment relevant to the query is less than 75\% of the original video length.
A system that can retrieve moments from videos would help people directly watch the video portion they are interested in and save time.
For the retrieved moments, we collect more fine-grained annotations by dividing the moment into steps and captioning each step. We explain the moment annotation below.

\vspace{3pt} \par \noindent\textbf{Stage 2: Moment Segmentation and Step Captions.}
In this stage, we collect fine-grained, stepwise annotations of the retrieved moments.
We ask crowdworkers to watch retrieved moments, divide them into several steps and mark the start timestamp of each step.
Then, for each of the marked moment segments, they are asked to write a \textit{step caption} that describes the specific step to complete (\eg ``add crayons to the candle'', ``melt it in bowl with hot water'', ``stir it well until dry'').
Our text queries from HowTo100M~\cite{miech19howto100m} are instructional questions starting with ``how to'', and we want the step captions to serve as short textual summaries of moments/steps.
we ask crowdworkers to start each caption with an action verb (\eg ``add", ``apply") and limit the length of the captions to seven words.

\begin{figure*}[t]
    \centering
    \includegraphics[width=.95\textwidth]{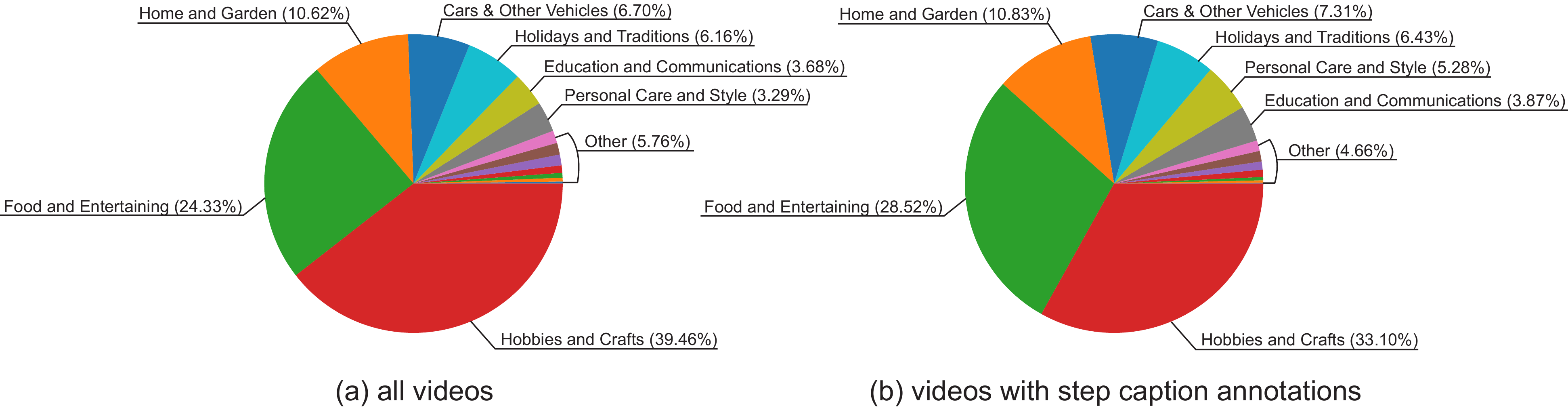}
    \caption{
        Task category distribution of \dataname{} text queries.
        There are a wide variety of categories for our videos.
        The most frequent categories are ``Hobbies and Crafts", ``Food and Entertaining", and ``Home and Garden". The task categories are from HowTo100M~\cite{miech19howto100m}.
    }
    \label{fig:datasetcategorydistribution}
\end{figure*}

\subsection{Dataset Analysis}
\label{sec:dataset_analysis}

\vspace{3pt} \par \noindent\textbf{Task Category Distribution.}
Our videos and text queries are collected from the HowTo100M~\cite{miech19howto100m} dataset, and hence our category labels match theirs. As shown in \cref{fig:datasetcategorydistribution}, the most frequently occurring categories (for all text-video pairs and just videos with step captions) are ``Hobbies and Crafts", ``Food and Entertaining", and ``Home and Garden". While these are the most common categories (similar to HowTo100M's most common categories), other categories still have a presence in our dataset. 

\vspace{3pt} \par
\begin{figure}[t]
    \centering
    \includegraphics[width=0.90\columnwidth]{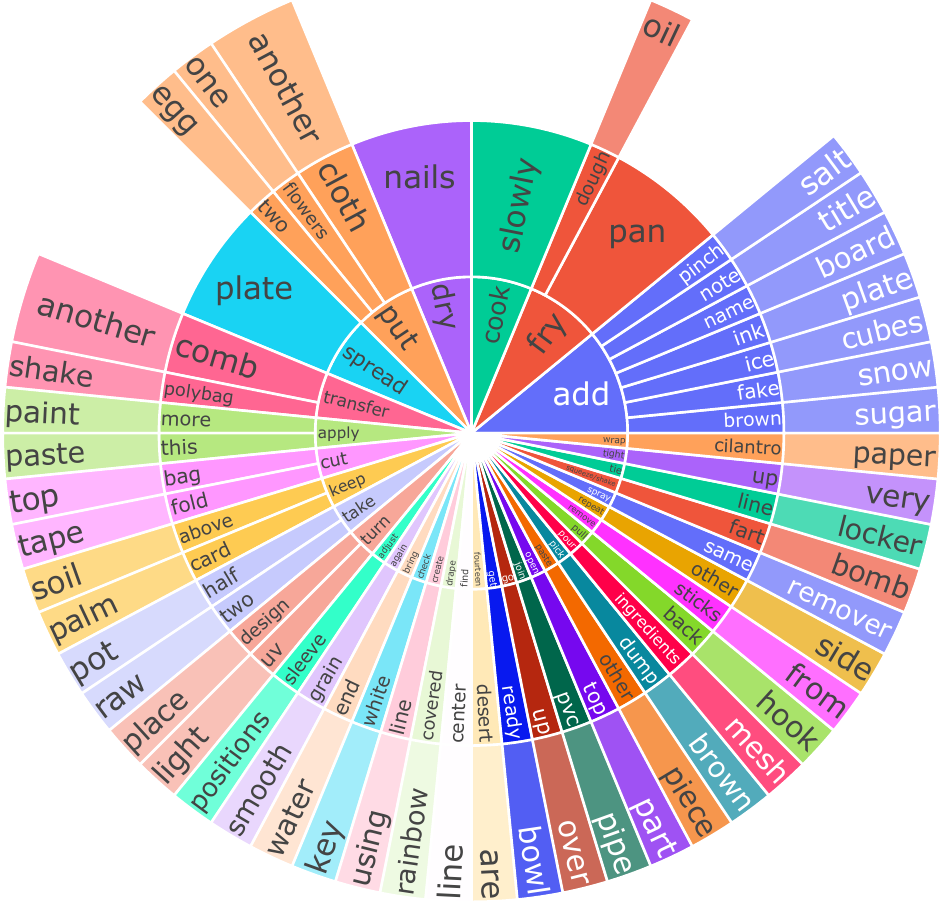}
    \caption{
        Distribution of \dataname{} step captions by their first three words for 50 random samples. Words are often related to actions or objects. We remove stop words (\eg `the', `it', \etc). 
    }
    \label{fig:step_caption_sunburst}
\end{figure}

\begin{figure}[t]
    \centering
    \includegraphics[width=0.95\columnwidth]{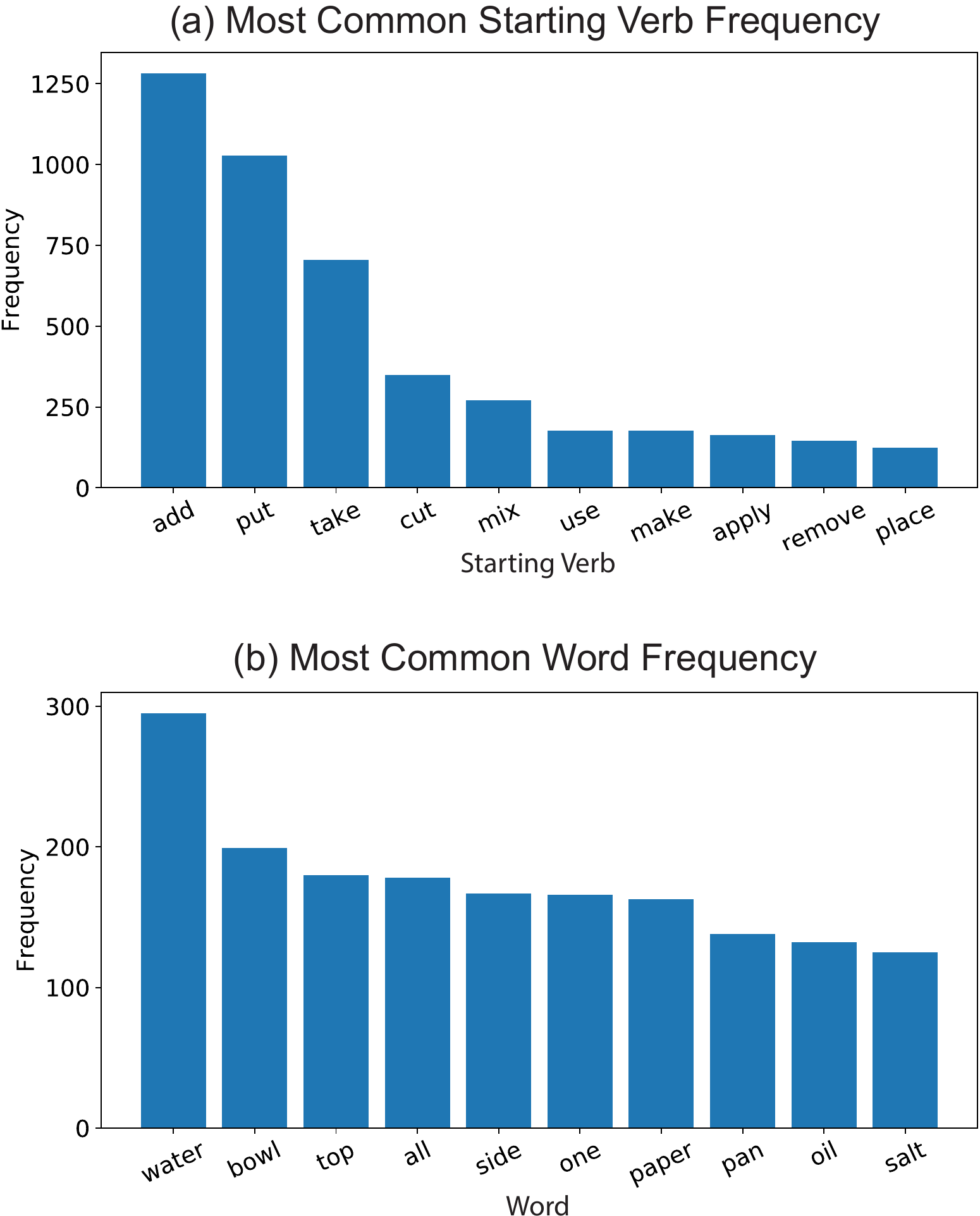}
    \caption{
        (a) Top 10 most common starting verbs in \dataname{} step captions.
        (b) Top 10 most common words in \dataname{} step captions (excluding the starting words and stop words). The top words typically refer to objects (\eg water) or quantities (\eg all).
    }
    \label{fig:top10words}
\end{figure}

 \noindent\textbf{Dataset Statistics.}
We collected a total of 3.4K text-video pairs, which are 287 seconds long on average, with a total duration of 270 hours.
Out of 3.4K videos, 1.8K videos are \textit{clippable} to a moment; \ie, only a short clip ($<$75\% of the original video) is relevant to the text query.
The average moment length is 148 seconds, which is 55\% of the original videos.
Out of the 1.8K moments, we provide moment segmentation and step caption annotations for the randomly chosen 1.1K moments.
The 1.1K moments are broken down to 7.6 steps on average, totaling 8.6K steps.
Each step is annotated with a start-end timestamp and a step caption.
The step captions are on average 4.42 words long and have 633 unique starting verbs with 3382 unique words.
\cref{fig:top10words} shows the most frequent starting verbs and the most frequent words in the step captions (not counting the starting word and stop words).
\cref{fig:step_caption_sunburst} shows the first three words of 50 random step caption samples (ignoring stop words).
As shown in the visualizations, the manually written step captions of \dataname{} cover open domain instruction steps and have a diverse vocabulary.

\vspace{3pt} \par \noindent\textbf{Comparisons to Other Datasets with Step Captions.}
\Cref{tbl:datasetcompare} compares our \dataname{} dataset to other video datasets with step annotations.
\dataname{} covers various open-domain videos with many step annotations per video and high-quality step captions written by human annotators.
While COIN~\cite{Tang2019COIN} and CrossTask \cite{Zhukov2019CrossTask} also provide step-level annotations for open-domain videos, however, they are restricted to a set of predefined steps.
In contrast, all the step captions of \dataname{} are manually written to answer the input text query.

\vspace{3pt} \par \noindent\textbf{Data Splits.}
Since there are cases where multiple videos are retrieved from the same query, we split our dataset into train/val/test splits by query instead of video.
We split our queries into 546/292/546 (1507/477/1391 videos) for train/val/test splits, respectively.

\subsection{Hierarchical Tasks Enabled by \dataname{}}
\label{sec:tasks}

In the following, we introduce four tasks connected in a hierarchy based on our \dataname{} dataset.
See \cref{fig:mainfigure} for an overview and visual examples of the tasks.

\vspace{3pt} \par \noindent\textbf{Video Retrieval.}
This task gives models an instructional text query (\eg ``How to make a memory jar''), and the models need to determine which videos are relevant and retrieve the top results. The models must retrieve videos among 
4.2K test split videos (1.4K videos paired with text queries + 2.8K distractor videos from HowTo100M~\cite{miech19howto100m}).
Distractor videos serve as negative examples (hence `distractors'), similar to Revaud \etal~\cite{Revaud2013CVPR}. We include these distractors to help increase the difficulty of our video retrieval task.

\vspace{3pt} \par \noindent\textbf{Moment Retrieval.}
In this task, the goal is to extract the portion of the video that is directly relevant to the given text query (i.e. to remove any unnecessary information from the start/end of the video).

\vspace{3pt} \par \noindent\textbf{Moment Segmentation.}
In this task, models should identify all relevant key `steps' from the retrieved relevant moment of the video. Models should generate a list of start and end times for every key step in a given video.

\vspace{3pt} \par \noindent\textbf{Step Captioning.}
This task requires models to generate short textual step captions for each retrieved step in a video. Models are provided with the source video and start/end times of each step. They should then generate a short instructional step caption for every step.

\section{Experiments}
\label{sec:Experiments}

For all four \dataname{} tasks,
we conduct experiments with
task-specific baseline models (\cref{sec:models}),
a joint baseline model  (\cref{sec:joint_model}),
and evaluate them with different standard metrics (\cref{sec:metrics}).
We represent each video as 32 frames with uniform intervals, if not specified.

\begin{figure*}[t]
    \centering
    \includegraphics[width=.90\linewidth]{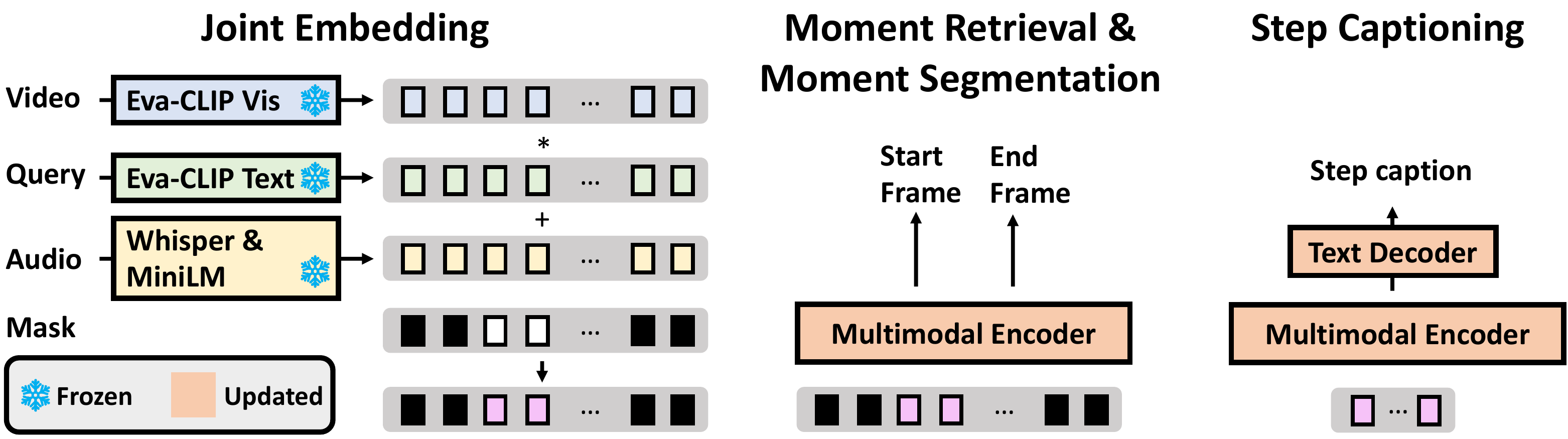}
    \caption{Illustration of our joint model  that handles moment retrieval, moment segmentation, and step captioning tasks (\cref{sec:joint_model}).
    We learn a shallow multimodal transformer encoder layer that adapts the four pretrained models:
    EVA-CLIP (frozen),
    Whisper (frozen),
    MiniLM (frozen),
    and CLIP4Caption (finetuned).
    }
    \label{fig:joint_model}
\end{figure*}

\subsection{Task-specific Models}
\label{sec:models}

\vspace{3pt} \par \noindent\textbf{Video Retrieval.}
We experiment with
CLIP (ViT-B/32)~\cite{Radford2021CLIP},
EVA-CLIP (ViT-G/14)~\cite{Fang2023EVA},
Frozen-in-Time~\cite{Bain21FrozenInTime},
and
MIL-NCE (S3D)~\cite{miech19endtoend},
which are pretrained text-to-image (CLIP/EVA-CLIP) and text-to-video (Frozen-in-Time/MIL-NCE) retrieval models, respectively.
For CLIP and EVA-CLIP,
we obtain a video embedding by averaging frame embeddings.
We compute the matching score by taking the cosine similarity between video and text query embedding.
Following the original setup, we use 4 frames for Frozen-in-Time and 32 frames for MIL-NCE.

\vspace{3pt} \par \noindent\textbf{Moment Retrieval.}
We experiment with two CLIP-based heuristics methods and the event proposal module of BMT~\cite{BMT_Iashin_2020}, a dense video captioning model pretrained on ActivityNet Captions~\cite{Krishna2017}.
With CLIP, we compute the cosine similarity between all frames and the text query and find the frame with the highest score.
Then we determine the start/end boundary of a moment with two different heuristics:
1) picking the frames where the similarity score drops from the highest scoring frame by a certain threshold (\eg, 0.10);
2) picking the 8 frames to the left and right, totaling up to 17 (= 8+1+8) frames
(see appendix for details).
Furthermore, we experiment with the BMT~\cite{BMT_Iashin_2020} event proposal module, which predicts video event proposals with center/length/confidence values.
We allow BMT to generate various events and then take the minimum start time and maximum end time across the events as the retrieved moment.
For BMT, we give the model the I3D~\cite{Carreira2017QuoVAI3D} RGB+Flow features and VGGish~\cite{Hershey2017CNNAFVGGish} audio features of the entire video, extracted at 1fps.

\vspace{3pt} \par \noindent\textbf{Moment Segmentation.}
We experiment with 1) framewise difference with the Structural Similarity Index Measure (SSIM)~\cite{Wang2004SSIM}, and
2) the event proposal module of BMT~\cite{BMT_Iashin_2020}.
For SSIM, if two adjacent frames have an SSIM below a certain threshold (\eg, 0.85), we mark that as a step boundary.
For BMT, we feed the model I3D and VGGish features (extracted at 1fps) of the entire video and directly use the video event proposal prediction.

\vspace{3pt} \par \noindent\textbf{Step Captioning.}
We experiment with BMT and SwinBERT~\cite{lin2021SwinBERTend-to-end}, a pretrained video captioning model.
For BMT, we use I3D and VGGish features of each step, extracted at 1fps.
We do not use its event proposal module for this task, as we give the features within the ground-truth step boundaries.
For SwinBERT, we use YouCook2~\cite{Zhou2018TowardsALYouCook2} checkpoint and 32 video frames from each step as input to the model.

\subsection{Joint Model}
\label{sec:joint_model}

We also experiment with an end-to-end joint baseline model that handles moment retrieval, moment segmentation, and step captioning tasks with a single architecture.
As shown in \cref{fig:joint_model}, our model 
is built on four existing pretrained models:
EVA-CLIP~\cite{Fang2023EVA},
Whisper~\cite{Radford2022whisper},
MiniLM~\cite{reimers-2019-sentence-bert},
and
CLIP4Caption~\cite{Tang2021CLIP4Caption}.
EVA-CLIP visual encoder maps a video frame into a visual embedding,
EVA-CLIP text encoder maps a text query into a text embedding,
Whisper extracts speech transcription from audio,
MiniLM text encoder maps the speech transcription into a text embedding.
To adapt the video, text, and audio embeddings,
we finetune a two-layer multimodal encoder and a two-layer text decoder, which are initialized from CLIP4Caption (MSRVTT~\cite{Xu2016MSRVTT} checkpoint).
We train the joint model in a multi-task setup in a round-robin fashion, by sampling a batch from one of the data loaders at each step~\cite{Cho2021}.

\vspace{3pt} \par \noindent\textbf{Input Embedding.}
We construct the multimodal input embedding to the transformer by combining
1) EVA-CLIP video frame embedding,
2) EVA-CLIP text query embedding (tiled to the number of video frames),
3) and MiniLM speech transcription embedding (temporally warped into each frame),
and
4) task-specific mask embeddings.
For moment retrieval and moment segmentation tasks, we feed the same multimodal embeddings while masking out the frames that are outside of interest.

\vspace{3pt} \par \noindent\textbf{Moment Retrieval \& Moment Segmentation.}
Following the span-based text question answering models~\cite{Seo2017, Devlin2019},
we learn linear layers that predict the boundaries of moments and steps.
Concretely, we use three linear layers predicting moment start, moment end, and step boundaries.
For the moment retrieval, our joint start and end predictor predicts the moment boundary in parallel, and we do not mask out the video inputs.
For the moment segmentation, our joint model autoregressively predicts each step's boundaries with masking; \ie, we mask out
1) frames that are outside of the moment
and 2) frames that are included in the previous steps.
For both tasks, we feed the video in 1fps.

\vspace{3pt} \par \noindent\textbf{Step Captioning.}
Following CLIP4Caption~\cite{Tang2021CLIP4Caption},
we sample 20 frames from each step.
The autoregressive text decoder attends to the multimodal encoder output via cross-attention and generates each step caption independently.

\begin{table}[t]
    \centering
    \resizebox{0.95\columnwidth}{!}{
    \begin{tabular}{l c c c c c}
        \toprule
        Model & Frames & FT & R@1 & R@5 & R@10 \\
        \midrule

        CLIP-B/32 & 1 & & 11.4 & 20.7 & 27.3 \\
        CLIP-B/32 & 4 & & 12.5 & 28.8 & 37.4 \\
        CLIP-B/32 & 10 & & 13.0 & 31.7 & 39.9 \\
        CLIP-B/32 & 20 & & 13.0 & 33.3 & 41.2 \\
        CLIP-B/32 & 32 & & 12.6 & 33.0 & 41.8 \\

        Frozen-in-Time & 4 & & 7.0 & 19.4 & 26.7 \\
        MIL-NCE (S3D) & 32 &  & 13.9 & 31.1 & 41.4 \\

        \midrule

        CLIP-B/32 & 1 & \cmark & 11.5 & 22.7 & 27.1 \\
        CLIP-B/32 & 4 & \cmark & 13.9 & 29.5 & 39.4 \\
        CLIP-B/32 & 10 & \cmark & 11.4 & 31.3 & 41.4 \\
        CLIP-B/32 & 20 & \cmark & 12.3 & 31.7 & 41.6 \\
        CLIP-B/32 & 32 & \cmark & 13.0 & 32.1 & 41.9 \\

        \midrule
        
        EVA-CLIP-G/14 & 1 & & 18.9 & 32.6 & 37.5 \\
        EVA-CLIP-G/14 & 4 & & 20.7 & 43.6 & 53.7 \\
        EVA-CLIP-G/14 & 10 & & 26.0 & 48.5 & 58.8\\
        EVA-CLIP-G/14 & 20 & & \textbf{26.4} & \textbf{51.1} & \textbf{61.5} \\
        EVA-CLIP-G/14 & 32 & & 26.0 & 50.0 & 61.4\\

        \bottomrule
    \end{tabular}
    }
    \caption{
        Video retrieval results on \dataname{} test split.
        CLIP/EVA-CLIP results are based on temporal average pooling.
        \textit{FT: finetuning on \dataname{}, R@k: Recall@k.}
        MIL-NCE was trained on the HowTo100M dataset, which is the video source of \dataname{}.
    }
    \label{tab:video_retrieval}
\end{table}

\begin{table}[t]
    \centering
    \resizebox{.95\columnwidth}{!}{
    \begin{tabular}{l c c c}
        \toprule
        Model  & FT &  R@0.5 & R@0.7 \\
        \midrule
        CLIP-B/32 (threshold=0.05) & & 21.01 & 9.02 \\
        CLIP-B/32 (8 frames left/right) & & 34.02 & 15.72 \\
        EVA-CLIP-G/14 (threshold=0.10) & & 19.33 & 7.86 \\
        EVA-CLIP-G/14 (8 frames left/right) & & 38.27 & 19.33 \\
        BMT & & 43.56 & 10.57 \\
        \midrule
        BMT & \cmark & 71.91 & \textbf{39.18} \\
        Joint (Ours) & \cmark & \textbf{73.32} & 32.60 \\
        
        \bottomrule
    \end{tabular}
    }
    \caption{
        Moment retrieval
        results on \dataname{} test split.
        CLIP (threshold): determines the start/end frames, by picking the frames where the similarity score drops from the highest scoring frame with a certain threshold (\eg, 0.05).
        CLIP (8 frames left/right): determines the start/end frames by eight frames to the left and to the right of the highest scoring frame.
        \textit{FT: Finetuning on \dataname{}, R@IoU: Recall@1 with a threshold of IoU}.
    }
    \label{tab:moment_retrieval}
\end{table}

\subsection{Metrics}
\label{sec:metrics}

\vspace{3pt} \par \noindent\textbf{Video Retrieval.}
Following previous work~\cite{lei2020tvr,Li2020Hero,Yu2017,Bain21FrozenInTime,Li2021VALUE},
We evaluate models on Recall@k metrics: R@1, R@5, and R@10.

\vspace{3pt} \par \noindent\textbf{Moment Retrieval.}
Following previous work~\cite{lei2020tvr,Lei2021QVHighlightsDM},
we evaluate model outputs against the ground-truth (GT) moment spans with Recall@1 with Intersection over Union (IoU) thresholds (0.5 and 0.7).

\vspace{3pt} \par \noindent\textbf{Moment Segmentation.}
Following previous work~\cite{lei2020tvr,Lei2021QVHighlightsDM},
we evaluate models on how similar the generated step spans are to the GT spans using IoU.
We then compute the recall and precision with IoU thresholds (0.5 and 0.7).

\vspace{3pt} \par \noindent\textbf{Step Captioning.}
Following previous work~\cite{Xu2016MSRVTT,lin2021SwinBERTend-to-end,BMT_Iashin_2020,Li2021VALUE},
we evaluate with the N-gram metrics:
CIDEr~\cite{Vedantam2015CIDErCI}, METEOR~\cite{Banerjee2005METEORAA}, and SPICE~\cite{Anderson2016SPICESP} with the language\_evaluation package.\footnote{\url{https://github.com/bckim92/language-evaluation}}
We also report two sentence-level embedding-based metrics  BERTScore~\cite{Zhang2019BertScore} and CLIPScore~\cite{Hessel2021}.

For BERTScore, we use the RoBERTa-Large~\cite{Liu2019c}.
For CLIPScore, we CLIP ViT-B/32~\cite{Radford2021CLIP} and report the average of frame-caption cosine similarities using 4 frames uniformly sampled from each step.
In addition, we compute the entailment of generated sentences to the GT sentences using the ELMo~\cite{Peters2018}-based Decomposable Attention model~\cite{Parikh2016ADAEntailment} pretrained on SNLI~\cite{Bowman2015} with 3 labels: \{\texttt{entailment}, \texttt{contradict}, \texttt{neutral}\}.\footnote{\url{https://docs.allennlp.org/models/main/models/pair_classification/models/decomposable_attention/}}
We use the ratio of \texttt{entailment} prediction as the entailment score.

\begin{table}[t]
    \centering
    \resizebox{0.99\columnwidth}{!}{
    \begin{tabular}{l c c c c c c}
        \toprule
        \multirow{2}{*}{Model} & \multirow{2}{*}{FT} & \multicolumn{2}{c}{Recall@IoU} & & \multicolumn{2}{c}{Precision@IoU} \\
        \cmidrule{3-4}
        \cmidrule{6-7}
        & & 0.5 & 0.7 & & 0.5 & 0.7 \\
        \midrule
        SSIM@0.75 (32 frames) & & 12.24 & 5.27 & & 26.32 & 10.05 \\
        SSIM@0.85 (32 frames) & & 25.03 & 9.79 & & \textbf{37.38} & \textbf{13.80} \\
        BMT (1fps) & & 8.24 & 3.71 & & 20.95 & 7.96 \\
        \midrule
        BMT (1 fps) & \cmark & 34.07 & 12.35 & & 24.71 & 8.93 \\
        Joint (Ours) (1 fps) & \cmark & \textbf{37.50} & \textbf{14.76} & & 28.52 & 10.84 \\
        \bottomrule
    \end{tabular}
    }
    \caption{
        Moment segmentation
        results on \dataname{} test split.
        We perform zeroshot evaluation with BMT, and then also provide results of using SSIM.
        SSIM is given 32 frames.
        \textit{FT: Finetuning on \dataname{}, Recall/Precision@IoU:  Recall@1/Precision with a threshold of IoU, SSIM@k: SSIM with a score threshold of k.}
    }
    \label{tab:moment_segmentation}
\end{table}

\begin{table}[t]
    \centering
    \resizebox{0.99\columnwidth}{!}{
    \begin{tabular}{l c c c c c c c}
        \toprule
        Model & FT & METEOR & CIDEr & SPICE & Entail. (\%) & BERT-S & CLIP-S \\
        \midrule
        BMT & & 2.23 & 1.04 & 1.41 & 1.17 & 0.83 & 0.21 \\
        SwinBERT & & 5.12 & 13.31 & 4.65 & 5.86 & 0.85 & \textbf{0.23} \\

        \midrule
        BMT & \cmark & 3.84 & 6.72 & 1.05 & 30.68 & 0.82 & 0.20 \\
        SwinBERT & \cmark & \textbf{5.94} & \textbf{24.66} & \textbf{6.67} & 35.09 & \textbf{0.86} & \textbf{0.23} \\
         Joint (Ours) & \cmark & 4.13 & 23.01 & 3.54 & \textbf{43.88} & \textbf{0.86} & \textbf{0.23} \\
        \bottomrule
    \end{tabular}
    }
    \caption{
        Step captioning
        results on \dataname{} test split.
        We finetune each model on \dataname{} and evaluate them on our test split.
        \textit{FT: Finetuning on \dataname{}, Entail: Entailment, BERT-S: BERTScore, CLIP-S: CLIPScore}.
    }
    \label{tab:step_captioning}
\end{table}

\begin{figure*}[t]
    \centering
    \includegraphics[width=0.95\textwidth]{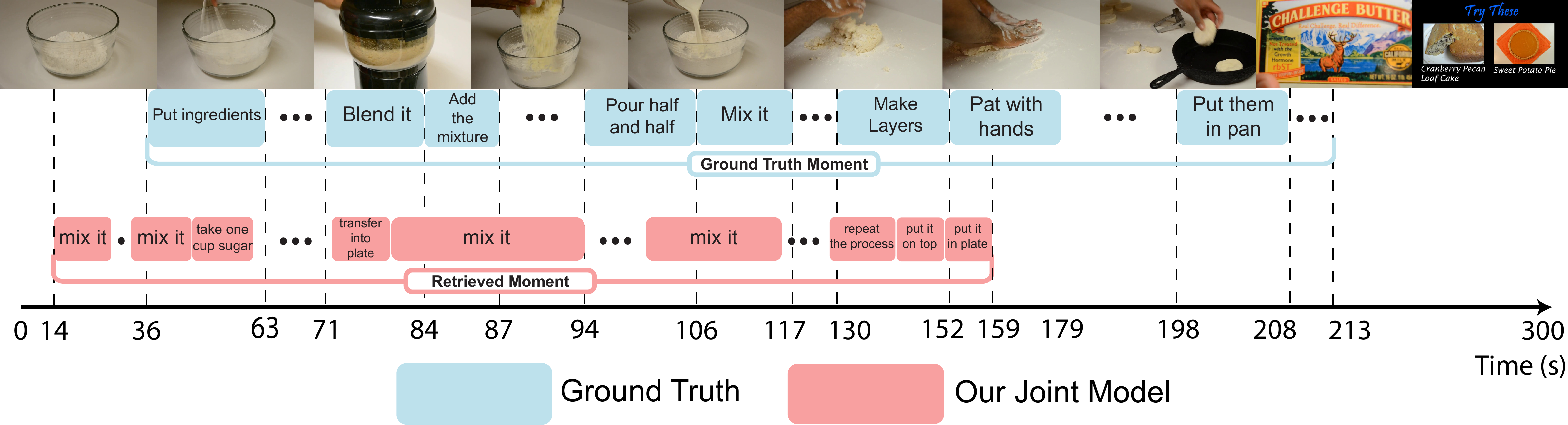}
    \caption{
        Comparison of our joint model prediction and ground truth annotation for moment retrieval, moment segmentation, and step captioning.
        The video is paired with a text query `How to make butter biscuits'.
    }
    \label{fig:modelvsgt_generationexample}
\end{figure*}

\section{Results and Discussions}
\label{sec:results}

In the following, we present the experiment results on the four tasks and the visualization of the pipelined model predictions.
Our baseline models show promising initial results, but there exists some gap between the current model performance and the upper bound accuracies, leaving large room for future improvements.

\vspace{3pt} \par \noindent\textbf{Video Retrieval.}
\Cref{tab:video_retrieval} shows the video retrieval results.
Increasing input frames increases the recall until 20 frames.
Although CLIP was not trained on a video dataset,
CLIP outperforms Frozen-in-Time (4 frames)
shows comparable performance with MIL-NCE (32 frames).
This is likely due to the fact that CLIP was trained on a much larger dataset than Frozen-in-Time.
Finetuning CLIP on \dataname{} does not show a big difference.
EVA-CLIP, a larger CLIP architecture with 1B parameters, outperforms all the other models with a big margin.
Thus, we use EVA-CLIP as our video retrieval model and use its features for the three downstream tasks for our joint model.

\vspace{3pt} \par \noindent\textbf{Moment Retrieval.}
\Cref{tab:moment_retrieval} shows the results for moment retrieval. 
Among the cosine similarity-based zero-shot methods, 
the 8-frame left/right method outperforms the similarity score drop difference method for both CLIP and EVA-CLIP.
BMT achieves better R@0.5 than the zero-shot methods,
and the finetuning improves both recall metrics.
Our joint model outperforms finetuned BMT on the R@0.5, while finetuned BMT achieves a higher score on R@0.7.

\vspace{3pt} \par \noindent\textbf{Moment Segmentation.}
Table~\ref{tab:moment_segmentation} shows the results for the moment segmentation task.
In the zero-shot setting, BMT fails to adapt the span distribution of \dataname{}, and simple SSIM methods could outperform the BMT model on both recall and precision.
But after finetuning, BMT shows significant improvement over its zero-shot version and SSIM methods 
on recall metrics.
Our joint model achieves a better performance than BMT on both recall and precision.

\vspace{3pt} \par \noindent\textbf{Step Captioning.}
Table~\ref{tab:step_captioning} shows the results of the step captioning task.
For both BMT and SwinBERT, zero-shot inference did not result in a good result in N-gram (\eg, CIDEr) and entailment metrics,
indicating the domain gap between their pretraining datasets (ActivityNet caption and YouCook2) and \dataname{} is not negligible. For example, their captions are longer than step captions of \dataname{}.
Finetuning brings a performance boost to BMT and SwinBERT in N-gram and entailment metrics but not in sentence-level embedding-based metrics (BERTScore and CLIPScore).
Compared to SwinBERT, our joint model achieves similar CIDEr and sentence-level embedding-based metrics. Notably, our joint model outperforms SwinBERT significantly on the entailment metric.
Future work on our dataset can also hopefully explore the complementary strengths of SwinBERT and our joint model.

\begin{table}[t]
    \centering
    \resizebox{0.99\columnwidth}{!}{
    \begin{tabular}{l c c c c c c c }
        \toprule

        \multirow{2}{*}{Model} & \multirow{2}{*}{FT} & \multicolumn{2}{c}{Moment Retrieval} & \multicolumn{2}{c}{Moment Segmentation} & Step Captioning \\
        \cmidrule(lr){3-4} \cmidrule(lr){5-6} \cmidrule(lr){7-7}
        & & R@0.5 & R@0.7 & R@0.7 & P@0.7 & CIDEr \\
        \midrule
        \multicolumn{7}{c}{With Audio} \\
        \midrule
        BMT & \cmark & 71.9 & 39.2 & 12.4 & 8.9 & 6.7 \\
        Joint (Ours) & \cmark & 73.3 & 32.6 & 14.8 & 10.8 & 23.0  \\
        \midrule
        \multicolumn{7}{c}{Without Audio} \\
        \midrule
        BMT & \cmark & 62.6 (\textcolor{red}{-9.3}) & 32.34 (\textcolor{red}{-6.8}) & 10.4 (\textcolor{red}{-2.0}) & 7.4 (\textcolor{red}{-1.6}) & 6.1 (\textcolor{red}{-0.6}) \\
        Joint (Ours) & \cmark & 70.7 (\textcolor{red}{-2.6}) & 20.6 (\textcolor{red}{-12.0}) & 13.5 (\textcolor{red}{-1.3}) & 10.0 (\textcolor{red}{-0.8}) & 15.2 (-\textcolor{red}{-7.8})  \\
        
        \bottomrule
        
    \end{tabular}
    }
    \caption{Ablation of using audio inputs.
    Removing audio input drops the performance of all three tasks for both models.
    }
    \label{tab:audio_ablation}
\end{table}

\vspace{3pt} \par \noindent\textbf{Audio Ablation.}
\Cref{tab:audio_ablation} shows the ablation study about using (top rows) and not using (bottom rows) audio input with BMT and our joint model.
Overall, both models show a performance drop without audio input.
For \momentretrievaltask{} and \stepretrievalsubtask{}, removing audio input significantly drops the scores for both models,
indicating that audio is very helpful for the tasks that require models to detect the boundaries of events.
For the \momentcaptioningtask{} task, removing audio input significantly drops the score for our joint model, while BMT does not show a big difference.

\vspace{3pt} \par \noindent\textbf{Visualization of Hierarchical Model Pipelining.}
In \cref{fig:modelvsgt_generationexample},
we visualize the model prediction results and ground-truth annotation for moment retrieval, moment segmentation, and step captioning tasks on a video associated with a query `How to make butter biscuits'.
The retrieved moment matches with the video moment about making the batter (36-159s) with the ground truth (GT) annotations.
The predicted step boundaries and step captions also show semantic correspondence with GT annotations and the video.
For example, the predicted caption `mix it'  matches the GT captions `add the mixture' (84-87s) and `mix it' (106-117s).
The model also captions `take one cup sugar' during that part where ingredients are added (47-55s).
The model makes mistakes by missing the end of the dough cutting and the final cooking process (160-213s) during moment retrieval. In this period, we find that a human instructor stands up and describes the process, making the frames visually very different from the previous batter-making process.

\section{Conclusion}

In this work, we present the \dataname{} dataset and propose a new benchmark that covers hierarchy in information retrieval and summarization from an instructional video corpus.
Our benchmark consists of four tasks: \videoretrievaltask{}, \momentretrievaltask{}, and our new \stepretrievalsubtask{} and \momentcaptioningtask{} tasks.
Different from existing video datasets with step captions, our \dataname{} provides unique, diverse, high-quality instruction steps with timestamps written by human annotators.
We provide comprehensive dataset analysis and present experiments with several task-specific and end-to-end joint baseline models for each task as starting points.
We hope that \dataname{} can foster future work on multimodal systems for holistic video information retrieval, summarization, and step-by-step reasoning.

\section*{Acknowledgments}
We thank the reviewers for their helpful comments. This work was supported by Meta AI, ARO Award W911NF2110220, DARPA KAIROS Grant FA8750-19-2-1004, and NSF-AI Engage Institute DRL-211263. The views, opinions, and/or findings contained in this article are those of the authors and not of the funding agency. For this work, the collection of data and the subsequent experiments were performed by the University of North Carolina at Chapel Hill, and not by Meta AI. As a result, both the data and code will be released by UNC Chapel Hill.

{\small
\bibliographystyle{ieee_fullname}
\bibliography{citations}

\begin{thebibliography}{10}\itemsep=-1pt

\bibitem{Anderson2016SPICESP}
Peter Anderson, Basura Fernando, Mark Johnson, and Stephen Gould.
\newblock Spice: Semantic propositional image caption evaluation.
\newblock In {\em ECCV}, 2016.

\bibitem{Bain21FrozenInTime}
Max Bain, Arsha Nagrani, G{\"u}l Varol, and Andrew Zisserman.
\newblock Frozen in time: A joint video and image encoder for end-to-end
  retrieval.
\newblock In {\em IEEE International Conference on Computer Vision}, 2021.

\bibitem{Banerjee2005METEORAA}
Satanjeev Banerjee and Alon Lavie.
\newblock Meteor: An automatic metric for mt evaluation with improved
  correlation with human judgments.
\newblock In {\em IEEvaluation@ACL}, 2005.

\bibitem{Bowman2015}
Samuel~R. Bowman, Gabor Angeli, Christopher Potts, and Christopher~D. Manning.
\newblock {A large annotated corpus for learning natural language inference}.
\newblock In {\em EMNLP}, 2015.

\bibitem{Carreira2017QuoVAI3D}
Jo{\~a}o Carreira and Andrew Zisserman.
\newblock Quo vadis, action recognition? a new model and the kinetics dataset.
\newblock {\em 2017 IEEE Conference on Computer Vision and Pattern Recognition
  (CVPR)}, pages 4724--4733, 2017.

\bibitem{Cho2021}
Jaemin Cho, Jie Lei, Hao Tan, and Mohit Bansal.
\newblock {Unifying Vision-and-Language Tasks via Text Generation}.
\newblock In {\em ICML}, feb 2021.

\bibitem{Devlin2019}
Jacob Devlin, Ming-Wei Chang, Kenton Lee, and Kristina Toutanova.
\newblock {BERT: Pre-training of Deep Bidirectional Transformers for Language
  Understanding}.
\newblock In {\em NAACL}, oct 2019.

\bibitem{Fang2023EVA}
Yuxin Fang, Wen Wang, Binhui Xie, Quan Sun, Ledell Wu, Xinggang Wang, Tiejun
  Huang, Xinlong Wang, and Yue Cao.
\newblock Eva: Exploring the limits of masked visual representation learning at
  scale.
\newblock In {\em CVPR}, 2023.

\bibitem{Gygli2014CreatingSF}
Michael Gygli, Helmut Grabner, Hayko Riemenschneider, and Luc~Van Gool.
\newblock Creating summaries from user videos.
\newblock In {\em ECCV}, 2014.

\bibitem{Hendricks2017LocalizingMI}
Lisa~Anne Hendricks, Oliver Wang, Eli Shechtman, Josef Sivic, Trevor Darrell,
  and Bryan~C. Russell.
\newblock Localizing moments in video with natural language.
\newblock In {\em ICCV}, pages 5804--5813, 2017.

\bibitem{Hershey2017CNNAFVGGish}
Shawn Hershey, Sourish Chaudhuri, Daniel P.~W. Ellis, Jort~F. Gemmeke, Aren
  Jansen, R.~Channing Moore, Manoj Plakal, Devin Platt, Rif~A. Saurous, Bryan
  Seybold, Malcolm Slaney, Ron~J. Weiss, and Kevin~W. Wilson.
\newblock Cnn architectures for large-scale audio classification.
\newblock {\em 2017 IEEE International Conference on Acoustics, Speech and
  Signal Processing (ICASSP)}, pages 131--135, 2017.

\bibitem{Hessel2021}
Jack Hessel, Ari Holtzman, Maxwell Forbes, Ronan~Le Bras, and Yejin Choi.
\newblock {CLIPScore: A Reference-free Evaluation Metric for Image Captioning}.
\newblock In {\em EMNLP}, 2021.

\bibitem{BMT_Iashin_2020}
Vladimir Iashin and Esa Rahtu.
\newblock A better use of audio-visual cues: Dense video captioning with
  bi-modal transformer.
\newblock In {\em British Machine Vision Conference (BMVC)}, 2020.

\bibitem{Krishna2017}
Ranjay Krishna, Kenji Hata, Frederic Ren, Li Fei-Fei, and Juan~Carlos Niebles.
\newblock {Dense-Captioning Events in Videos}.
\newblock In {\em ICCV}, 2017.

\bibitem{Kuehne2014TheLOBreakfast}
Hilde Kuehne, Ali~Bilgin Arslan, and Thomas Serre.
\newblock The language of actions: Recovering the syntax and semantics of
  goal-directed human activities.
\newblock {\em 2014 IEEE Conference on Computer Vision and Pattern
  Recognition}, pages 780--787, 2014.

\bibitem{Lei2021QVHighlightsDM}
Jie Lei, Tamara~L. Berg, and Mohit Bansal.
\newblock Qvhighlights: Detecting moments and highlights in videos via natural
  language queries.
\newblock In {\em NeurIPS}, 2021.

\bibitem{lei2020tvr}
Jie Lei, Licheng Yu, Tamara~L Berg, and Mohit Bansal.
\newblock Tvr: A large-scale dataset for video-subtitle moment retrieval.
\newblock In {\em ECCV}, 2020.

\bibitem{Li2020Hero}
Linjie Li, Yen-Chun Chen, Zhe~Gan {Yu Cheng}, Licheng Yu, and Jingjing Liu.
\newblock {HERO: Hierarchical Encoder for Video+Language Omni-representation
  Pre-training}.
\newblock In {\em EMNLP}, 2020.

\bibitem{Li2021VALUE}
Linjie Li, Jie Lei, Zhe Gan, Licheng Yu, Yen-Chun Chen, Rohit Pillai, Yu Cheng,
  Luowei Zhou, Xin~Eric Wang, William~Yang Wang, Tamara~Lee Berg, Mohit Bansal,
  Jingjing Liu, Lijuan Wang, and Zicheng Liu.
\newblock {VALUE: A Multi-Task Benchmark for Video-and-Language Understanding
  Evaluation}.
\newblock In {\em NeurIPS}, pages 1--21, 2021.

\bibitem{lin2021SwinBERTend-to-end}
Kevin Lin, Linjie Li, Chung-Ching Lin, Faisal Ahmed, Zhe Gan, Zicheng Liu,
  Yumao Lu, and Lijuan Wang.
\newblock Swinbert: End-to-end transformers with sparse attention for video
  captioning.
\newblock In {\em CVPR}, 2022.

\bibitem{Liu2019c}
Yinhan Liu, Myle Ott, Naman Goyal, Jingfei Du, Mandar Joshi, Danqi Chen, Omer
  Levy, Mike Lewis, Luke Zettlemoyer, and Veselin Stoyanov.
\newblock {RoBERTa: A Robustly Optimized BERT Pretraining Approach}, 2019.

\bibitem{miech19endtoend}
Antoine Miech, Jean-Baptiste Alayrac, Lucas Smaira, Ivan Laptev, Josef Sivic,
  and Andrew Zisserman.
\newblock End-to-end learning of visual representations from uncurated
  instructional videos.
\newblock In {\em CVPR}, 2020.

\bibitem{miech19howto100m}
Antoine Miech, Dimitri Zhukov, Jean-Baptiste Alayrac, Makarand Tapaswi, Ivan
  Laptev, and Josef Sivic.
\newblock How{T}o100{M}: {L}earning a {T}ext-{V}ideo {E}mbedding by {W}atching
  {H}undred {M}illion {N}arrated {V}ideo {C}lips.
\newblock In {\em ICCV}, 2019.

\bibitem{Narasimhan2022TLDWSI}
Medhini Narasimhan, Arsha Nagrani, Chen Sun, Michael Rubinstein, Trevor
  Darrell, Anna Rohrbach, and Cordelia Schmid.
\newblock Tl;dw? summarizing instructional videos with task relevance \&
  cross-modal saliency.
\newblock In {\em ECCV}, volume abs/2208.06773, 2022.

\bibitem{Parikh2016ADAEntailment}
Ankur~P. Parikh, Oscar T{\"a}ckstr{\"o}m, Dipanjan Das, and Jakob Uszkoreit.
\newblock A decomposable attention model for natural language inference.
\newblock In {\em EMNLP}, 2016.

\bibitem{Peters2018}
Matthew~E. Peters, Mark Neumann, Mohit Iyyer, Matt Gardner, Christopher Clark,
  Kenton Lee, and Luke Zettlemoyer.
\newblock {Deep contextualized word representations}.
\newblock In {\em NAACL}, 2018.

\bibitem{Radford2021CLIP}
Alec Radford, Jong~Wook Kim, Chris Hallacy, Aditya Ramesh, Gabriel Goh,
  Sandhini Agarwal, Girish Sastry, Amanda Askell, Pamela Mishkin, Jack Clark,
  Gretchen Krueger, and Ilya Sutskever.
\newblock Learning transferable visual models from natural language
  supervision.
\newblock {\em ArXiv}, abs/2103.00020, 2021.

\bibitem{Radford2022whisper}
Alec Radford, Jong~Wook Kim, Tao Xu, Greg Brockman, Christine McLeavey, and
  Ilya Sutskever.
\newblock Robust speech recognition via large-scale weak supervision, 2022.

\bibitem{reimers-2019-sentence-bert}
Nils Reimers and Iryna Gurevych.
\newblock Sentence-bert: Sentence embeddings using siamese bert-networks.
\newblock In {\em Proceedings of the 2019 Conference on Empirical Methods in
  Natural Language Processing}. Association for Computational Linguistics, 11
  2019.

\bibitem{Revaud2013CVPR}
Jérôme Revaud, Matthijs Douze, Cordelia Schmid, and Hervé Jégou.
\newblock Event retrieval in large video collections with circulant temporal
  encoding.
\newblock In {\em CVPR}, 2013.

\bibitem{Rohrbach2012MP2}
Marcus Rohrbach, Sikandar Amin, Mykhaylo Andriluka, and Bernt Schiele.
\newblock A database for fine grained activity detection of cooking activities.
\newblock In {\em 2012 IEEE Conference on Computer Vision and Pattern
  Recognition}, pages 1194--1201, 2012.

\bibitem{Seo2017}
Minjoon Seo, Aniruddha Kembhavi, Ali Farhadi, and Hananneh Hajishirzi.
\newblock {Bi-Directional Attention Flow for Machine Comprehension}.
\newblock In {\em ICLR}, pages 1--12, 2017.

\bibitem{Sharghi2017QueryFocusedVS}
Aidean Sharghi, Jacob~S. Laurel, and Boqing Gong.
\newblock Query-focused video summarization: Dataset, evaluation, and a memory
  network based approach.
\newblock In {\em CVPR}, pages 2127--2136, 2017.

\bibitem{Song2015TVSumSW}
Yale Song, Jordi Vallmitjana, Amanda Stent, and Alejandro Jaimes.
\newblock Tvsum: Summarizing web videos using titles.
\newblock In {\em CVPR}, pages 5179--5187, 2015.

\bibitem{Tang2021CLIP4Caption}
Mingkang Tang, Zhanyu Wang, Zhenhua Liu, Fengyun Rao, DIan Li, and Xiu Li.
\newblock {CLIP4Caption: CLIP for Video Caption}.
\newblock In {\em ACM MM}, 2021.

\bibitem{Tang2019COIN}
Yansong Tang, Dajun Ding, Yongming Rao, Yu Zheng, Danyang Zhang, Lili Zhao,
  Jiwen Lu, and Jie Zhou.
\newblock Coin: A large-scale dataset for comprehensive instructional video
  analysis.
\newblock In {\em IEEE Conference on Computer Vision and Pattern Recognition
  (CVPR)}, 2019.

\bibitem{Vedantam2015CIDErCI}
Ramakrishna Vedantam, C.~Lawrence Zitnick, and Devi Parikh.
\newblock Cider: Consensus-based image description evaluation.
\newblock {\em 2015 IEEE Conference on Computer Vision and Pattern Recognition
  (CVPR)}, pages 4566--4575, 2015.

\bibitem{wang-etal-2019-youmakeup}
Weiying Wang, Yongcheng Wang, Shizhe Chen, and Qin Jin.
\newblock {Y}ou{M}akeup: A large-scale domain-specific multimodal dataset for
  fine-grained semantic comprehension.
\newblock In {\em Proceedings of the 2019 Conference on Empirical Methods in
  Natural Language Processing and the 9th International Joint Conference on
  Natural Language Processing (EMNLP-IJCNLP)}, pages 5133--5143, Hong Kong,
  China, Nov. 2019. Association for Computational Linguistics.

\bibitem{Wang2004SSIM}
Zhou Wang, A.C. Bovik, H.R. Sheikh, and E.P. Simoncelli.
\newblock Image quality assessment: from error visibility to structural
  similarity.
\newblock {\em IEEE Transactions on Image Processing}, 13(4):600--612, 2004.

\bibitem{Wei2022COT}
Jason Wei, Xuezhi Wang, Dale Schuurmans, Maarten Bosma, Brian Ichter, Fei Xia,
  Ed Chi, Quoc Le, and Denny Zhou.
\newblock Chain-of-thought prompting elicits reasoning in large language
  models, 2022.

\bibitem{Xu2016MSRVTT}
Jun Xu, Tao Mei, Ting Yao, and Yong Rui.
\newblock {MSR-VTT: A Large Video Description Dataset for Bridging Video and
  Language}.
\newblock In {\em CVPR}, 2016.

\bibitem{Yu2017}
Youngjae Yu, Hyungjin Ko, Jongwook Choi, and Gunhee Kim.
\newblock {End-to-end concept word detection for video captioning, retrieval,
  and question answering}.
\newblock In {\em CVPR}, volume 2017-Janua, pages 3261--3269, 2017.

\bibitem{Zhang2019BertScore}
Tianyi Zhang, Varsha Kishore, Felix Wu, Kilian~Q. Weinberger, and Yoav Artzi.
\newblock {BERTScore: Evaluating Text Generation with BERT}.
\newblock In {\em ICLR}, 2019.

\bibitem{Zhang2023multimodalCOT}
Zhuosheng Zhang, Aston Zhang, Mu Li, Hai Zhao, George Karypis, and Alex Smola.
\newblock Multimodal chain-of-thought reasoning in language models, 2023.

\bibitem{Zhou2018TowardsALYouCook2}
Luowei Zhou, Chenliang Xu, and Jason~J. Corso.
\newblock Towards automatic learning of procedures from web instructional
  videos.
\newblock In {\em AAAI}, 2018.

\bibitem{Zhukov2019CrossTask}
Dimitri Zhukov, Jean-Baptiste Alayrac, Ramazan~Gokberk Cinbis, David Fouhey,
  Ivan Laptev, and Josef Sivic.
\newblock Cross-task weakly supervised learning from instructional videos.
\newblock In {\em 2019 IEEE/CVF Conference on Computer Vision and Pattern
  Recognition (CVPR)}, pages 3532--3540, 2019.

\end{thebibliography}
}


\appendix

In the appendix, we include the following content:
Data annotation details (\cref{sec:data_annotation_details}),
CLIP-based moment retrieval method visualization (\cref{sec:clip_similarity_method_detail}),
Model performance analysis in video categories and duration groups (\cref{sec:category_duration_analysis}),
and 
Evaluation details (\cref{sec:evaluation_detail}).

\section{Data Annotation Details}
\label{sec:data_annotation_details}

\paragraph{Annotation Interface.}

\begin{figure}[ht]
    \centering
    \includegraphics[width=.85\columnwidth]{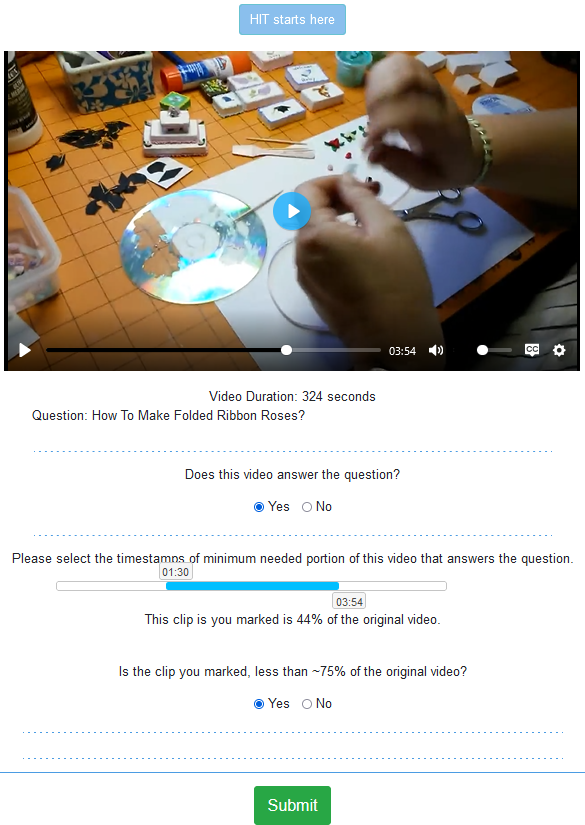}
    \caption{
        Stage 1 data collection interface for video and moment retrieval. Crowdworkers are presented with a video and text query and asked if the video answers/solves the question. If they select yes, the interface expands to a sliding bar that allows them to trim the video down to just the relevant portion.
    }
    \label{fig:stage1collectioninterface}
\end{figure}

\begin{figure}[ht]
    \centering
    \includegraphics[width=.85\columnwidth]{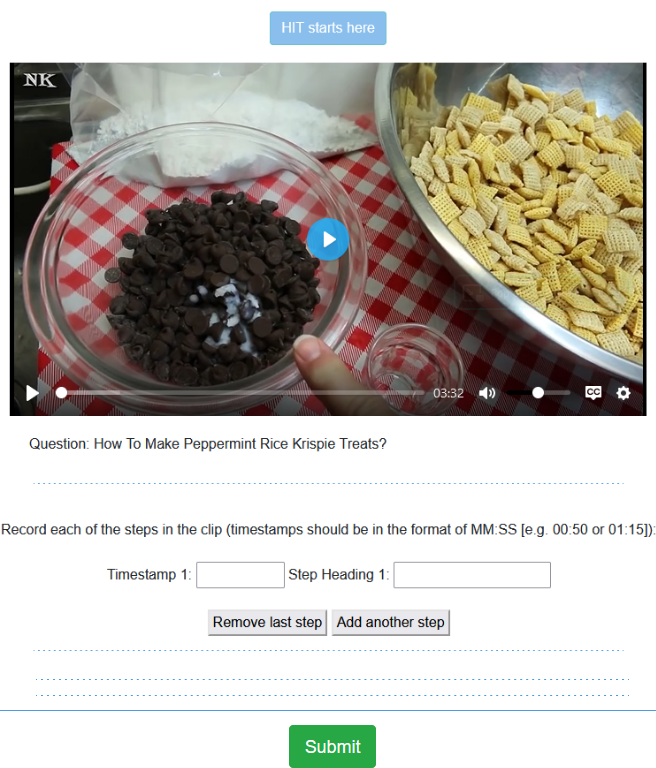}
    \caption{
        Stage 2 data collection interface moment segmentation and step captions. Crowdworkers are presented with a video and instructional text query and are asked to write all the essential steps in the video along with the timestamp of each step.
    }
    \label{fig:stage2collectioninterface}
\end{figure}

In the following, we provide screenshots of the \dataname{} annotation interface for each stage (\cref{fig:stage1collectioninterface} and \cref{fig:stage2collectioninterface})
and worker qualification process.

\paragraph{Worker Qualifications and Pay.}

We require crowdworkers to have above a 95\% approval rate and have completed at least 1000 or more other tasks before working on ours. We also require that all workers pass a qualification test (separate tests for each stage) before they can work on our tasks. 
For stage 1, the qualification test is composed of 2 parts. First workers are asked to determine if a video solves/answers the given prompt, then they are shown a relevant video and asked to identify the relevant moment in the video (we provide some leniency with timing).
For stage 2, workers are given a short series of videos with multiple-choice questions. The multiple choice answers consist of pre-written step captions for the video.
The workers were asked then to identify which set of step captions was the best (\ie best covered the video and didn't violate simple rules). A total of 72 workers passed the qualification test for both tasks.
As text queries from the HowTo100M~\cite{miech19howto100m} dataset are all in English and all of our collected step captions are in English, we require crowdworkers to be from an English-speaking country. Workers were paid \$0.20 for stage 1 and \$0.45 for stage 2. We also provide a large bonus for good workers. For stage 1, workers are bonused with \$0.05 if the video answers/solves the prompt and if they correctly trim the video.
Then for every 25 tasks they complete, their base pay increases by \$0.02. A typical worker can earn up to \$0.27 per task, which is roughly \$12.00 per hour. For stage 2, workers are paid with \$0.04 bonus for every high-quality step caption they write complete and for every 10 high-quality tasks they complete, their base pay is increased by \$0.02. A typical worker can earn up to \$0.81 per task, which is roughly \$12.00 per hour. For both tasks, there is a baseline pay of around \$12, but oftentimes, workers would complete more than 25 and 10 tasks (respectively, for stages 1 and 2), pushing the pay/hour higher.

\begin{figure}[t]
    \centering
    \includegraphics[width=\columnwidth]{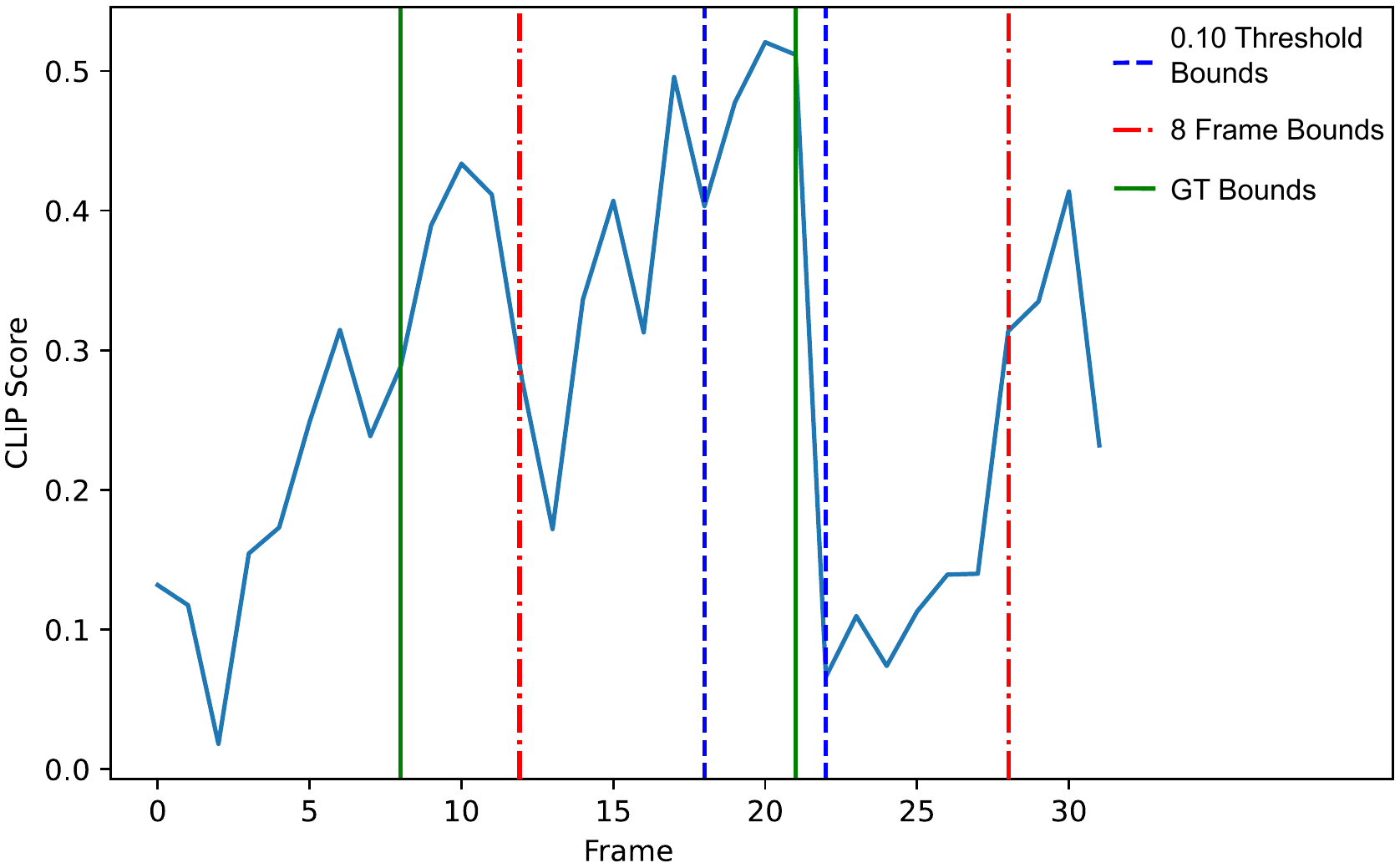}
    \caption{
        Visualization of image-text cosine similarity-based methods for the \momentretrievaltask{} task.
        The 8-frame method (\textcolor{red}{red} with dashes and dots) achieves IoU=0.42 while the 0.10 threshold method (\textcolor{blue}{blue} with dashes) achieves IoU=0.19.
        Ground Truth bounds are indicated with the solid \textcolor{green}{green} lines.
        EVA-CLIP model was used for the plot.
        \textit{CLIP Score: cosine similarity between image and text embedding}.
    }
    \label{fig:clipsimilarityplot}
\end{figure}

\section{CLIP-based Moment Retrieval Method}
\label{sec:clip_similarity_method_detail}

In \cref{fig:clipsimilarityplot}, we illustrate two heuristics that we discuss in the main paper \cref{sec:models}.
From the frame that scores the highest text-frame cosine similarity, we determine the start/end timestamp of the moment by
1) picking the frames where the similarity score drops from the highest scoring frame by a certain threshold (\eg, 0.10);
2) picking the 8 frames to the left and right, totaling up to 17 (= 8+1+8) frames.

\begin{table}[t]
    \centering
    \resizebox{.9\columnwidth}{!}{
    \begin{tabular}{l c c c}
        \toprule
        Category & \# Prompts &  \# Videos \\
        \midrule
        Hobbies and Crafts & 193 & 231 \\
        Food and Entertaining & 192 & 250 \\
        Home and Garden & 69 & 111 \\
        Cars and Other Vehicles & 28 & 55 \\
        Holidays and Traditions & 25 & 47 \\
        Education and Communications & 15 & 23 \\
        Personal Care and Style & 6 & 29 \\
        Pets and Animals & 5 & 6 \\
        Health & 5 & 13 \\
        Family Life & 4 & 1 \\
        Arts and Entertainment & 1 & 1 \\
        Sports and Fitness & 1 & 8 \\
        Misc. & 2 & 1 \\
        \midrule
        All & 546 & 776 \\
        \bottomrule
    \end{tabular}
    }
    \caption{
        Prompt and Video category distributions of \dataname{} test split.
        Categories are sorted in descending order by the number of prompts.
        The number of prompts is smaller than the number of videos since multiple videos were retrieved and paired with some prompts.
    }
    \label{tab:category_distributions}
\end{table}

\section{Model Performance Analysis in Video Categories and Duration}
\label{sec:category_duration_analysis}

As mentioned in the main paper \cref{sec:dataset}, \dataname{} videos are collected by text queries with different categories such as `Home and Garden' and `Food and Entertaining'. Also, videos and moments have different durations.
In \Cref{tab:category_distributions}, we show the distribution of prompts and videos for each category in \dataname{} test split.
In the following, we provide comprehensive evaluation results per category and different video/moment duration groups.

\begin{table}[t]
    \centering
    \resizebox{\columnwidth}{!}{
    \begin{tabular}{l c c c c c}
        \toprule
        Category & Model & FT & R@1 & R@5 & R@10 \\

        \midrule

        Hobbies and Crafts & EVA-CLIP & & 26.42 & 52.85 & 63.73 \\
        Food and Entertaining & EVA-CLIP & & 25.52 & 43.75 & 53.12 \\
        Home and Garden & EVA-CLIP & & 27.54 & 62.32 & 71.01 \\
        Cars and Other Vehicles & EVA-CLIP & & 14.29 & 53.57 & 64.29 \\
        Holidays and Traditions & EVA-CLIP & & 44.0 & 68.0 & 76.0 \\
        Education and Communications & EVA-CLIP & & 20.0 & 26.67 & 46.67 \\
        Personal Care and Style & EVA-CLIP & & 50.0 & 66.67 & 66.67 \\
        Pets and Animals & EVA-CLIP & & 0 & 60.0 & 80.0 \\
        Health & EVA-CLIP & & 60.0 & 60.0 & 80.0 \\
        Family Life & EVA-CLIP & & 0 & 75.0 & 100.0 \\
        Arts and Entertainment & EVA-CLIP & & 0 & 0 & 0 \\
        Sports and Fitness & EVA-CLIP & & 100.0 & 100.0 & 100.0 \\
        \midrule
        All & EVA-CLIP & & 26.37 & 51.1 & 61.54 \\
        \bottomrule
    \end{tabular}
    }
    \caption{
        Video retrieval results per prompt category on our \dataname{} test split.
        \textit{FT: finetuning on \dataname{}, R@k: Recall@k.}
    }
    \label{tab:video_retrieval_categories}
\end{table}

\paragraph{\Videoretrievaltask{}.}
In \Cref{tab:video_retrieval_categories},
we show
EVA-CLIP~\cite{Fang2023EVA} (ViT-G/14) with 20 frames
on each prompt category in our dataset.
Among the categories that have many most videos ($>$ 20 videos), the model is better at `Holidays and Traditions' and `Personal Care and Style' than `Cars and Other Vehicles'.

\begin{table}[t]
    \centering
    \resizebox{\columnwidth}{!}{
    \begin{tabular}{l c c c c}
        \toprule
        Category & Model & FT &  R@0.5 & R@0.7 \\
        \midrule
        Hobbies and Crafts & BMT & & 50.65 & 11.26  \\
        Food and Entertaining & BMT & & 34.40 & 8.80  \\
        Home and Garden & BMT & & 39.64 & 6.31  \\
        Cars and Other Vehicles & BMT & & 60.00 & 14.55\\
        Holidays and Traditions & BMT & & 36.17 & 6.38\\
        Education and Communications & BMT & & 47.83 & 26.09  \\
        Personal Care and Style & BMT & & 44.83 & 10.34  \\
        Pets and Animals & BMT & & 33.33 & 33.33  \\
        Health & BMT & & 61.54 & 30.77 \\
        Family Life & BMT & & 0 & 0 \\
        Arts and Entertainment & BMT & & 100 & 0 \\
        Sports and Fitness & BMT & & 62.5 & 12.5 \\
        \midrule
        All & BMT & & 43.56 & 10.57\\
        \midrule
        \midrule
        Hobbies and Crafts & BMT & \cmark & 72.29 & 39.39\\
        Food and Entertaining & BMT & \cmark & 72.80 & 38.00 \\
        Home and Garden & BMT & \cmark & 67.57 & 36.04 \\
        Cars and Other Vehicles & BMT & \cmark & 74.55 & 52.72 \\
        Holidays and Traditions & BMT & \cmark & 72.34 & 31.91 \\
        Education and Communications & BMT & \cmark & 60.87 & 39.13\\
        Personal Care and Style & BMT & \cmark & 72.41 & 34.38 \\
        Pets and Animals & BMT & \cmark & 66.67 & 16.67 \\
        Health & BMT & \cmark & 84.62 & 69.23  \\
        Family Life & BMT & \cmark & 100 & 100 \\
        Arts and Entertainment & BMT & \cmark & 100 & 100 \\
        Sports and Fitness & BMT & \cmark & 87.5 & 37.5 \\
        \midrule
        Hobbies and Crafts & Joint (Ours) & \cmark & 75.76 & 35.5 \\ 
        Food and Entertaining & Joint (Ours) & \cmark & 75.2 & 36.4 \\ 
        Home and Garden & Joint (Ours) & \cmark & 63.06 & 21.62 \\ 
        Cars and Other Vehicles & Joint (Ours) & \cmark & 81.82 & 34.55 \\ 
        Holidays and Traditions & Joint (Ours) & \cmark & 72.34 & 31.91 \\
        Education and Communications & Joint (Ours) & \cmark & 78.26 & 26.09 \\ 
        Personal Care and Style & Joint (Ours) & \cmark & 68.97 & 17.24 \\
        Pets and Animals & Joint (Ours) & \cmark & 33.33 & 33.33 \\   
        Health & Joint (Ours) & \cmark & 61.54 & 30.77 \\ 
        Family Life & Joint (Ours) & \cmark & 100.0 & 100.0 \\ 
        Arts and Entertainment & Joint (Ours) & \cmark & 100.0 & 0.0 \\ 
        Sports and Fitness & Joint (Ours) & \cmark & 75.0 & 50.0 \\ 
        \midrule
        All & BMT & \cmark & 71.91 & 39.18 \\
        All & Joint (Ours) & \cmark & 73.32 & 32.60 \\

        \bottomrule
    \end{tabular}
    }
    \caption{
        \Momentretrievaltask{} results per video category on our \dataname{} test split.
        \textit{FT: Finetuning on \dataname{}, R@IoU: Recall@1 with a threshold of IoU}.
    }
    \label{tab:moment_retrieval_categories}
\end{table}

\begin{table}[t]
    \centering
    \resizebox{.9\columnwidth}{!}{
    \begin{tabular}{l c c c c}
        \toprule
        Video Duration & Model & FT &  R@0.5 & R@0.7 \\
        \midrule
        $<$ 2 mins & BMT & & 48.70 & 10.43\\
        2 - 6 mins & BMT & & 37.47 & 7.04 \\
        $>$ 6 mins & BMT & & 56.74 & 20.22\\
        \midrule
        All & BMT & & 43.56 & 10.57\\
        \midrule
        \midrule
        $<$ 2 mins & BMT & \cmark & 74.78 & 44.35 \\
        2 - 6 mins & BMT & \cmark & 72.05 & 40.37 \\
        $>$ 6 mins & BMT & \cmark & 69.66 & 32.58 \\
        \midrule
        $<$ 2 mins & Joint (Ours) & \cmark & 68.10 & 18.10 \\
        2 - 6 mins & Joint (Ours) & \cmark & 73.21 & 28.21 \\
        $>$ 6 mins & Joint (Ours) & \cmark & 75.00 & 40.26 \\
        \midrule
        All & BMT & \cmark & 71.91 & 39.18 \\
        All & Joint (Ours) & \cmark & 73.32 & 32.60 \\
        \bottomrule
    \end{tabular}
    }
    \caption{
        \Momentretrievaltask{} results for various durations on our \dataname{} test split.
        \textit{FT: Finetuning on \dataname{}, R@IoU: Recall@1 with a threshold of IoU}.
    }
    \label{tab:moment_retrieval_durations}
\end{table}

\paragraph{\Momentretrievaltask{}.}
In \Cref{tab:moment_retrieval_categories}, we show the zeroshot and finetuning results of BMT~\cite{BMT_Iashin_2020} proposal module
and our joint model
on each \dataname{} video category.
Categories like `Home and Garden' and `Holidays and Traditions' see a strong performance increase after finetuning.

In \Cref{tab:moment_retrieval_durations}, we show \Momentretrievaltask{} performance on three video duration groups.
Before finetuning, BMT performs slightly better on videos of a longer length; however, after finetuning, BMT performs much better on shorter-length videos.
In R@0.5, our joint model outperforms BMT when the videos are longer.

\begin{table}[t]
    \centering
    \resizebox{\columnwidth}{!}{
    \begin{tabular}{l c c c c c c c}
        \toprule
        \multirow{2}{*}{Category} & \multirow{2}{*}{Model} & \multirow{2}{*}{FT} & \multicolumn{2}{c}{Recall@IoU} & & \multicolumn{2}{c}{Precision@IoU}\\
        \cmidrule{4-5}
        \cmidrule{7-8}
        & & & 0.5 & 0.7 & & 0.5 & 0.7 \\       
        \midrule
        Hobbies and Crafts & BMT & & 8.91 & 3.02 & & 22.44 & 6.17 \\
        Food and Entertaining & BMT & & 7.47 & 4.47 & & 19.06 & 9.36 \\
        Home and Garden & BMT & & 8.64 & 3.18 & & 23.33 & 7.62 \\
        Cars and Other Vehicles & BMT & & 4.18 & 1.11 & & 16.16 & 5.05 \\
        Holidays and Traditions & BMT & & 7.24 & 4.67 & & 16.11 & 10.50\\
        Education and Communications & BMT & & 6.11 & 0 & & 20.00 & 0 \\
        Personal Care and Style & BMT & & 9.87 & 6.75 & & 19.65 & 11.58 \\
        Pets and Animals & BMT & & 40.00 & 10.00 & & 80.00 & 20.00 \\
        Health & BMT & & 13.33 & 10.00 & & 30.00 & 20.00 \\
        Family Life & BMT & & 0 & 0 & & 0 & 0 \\
        Arts and Entertainment & BMT & & 0 & 0 & & 0 & 0 \\
        Sports and Fitness & BMT & & 0 & 0 & & 0 & 0 \\
        \midrule
        All & BMT & & 8.24 & 3.71 & & 20.95 & 7.96  \\
        \midrule
        \midrule
        Hobbies and Crafts & BMT & \cmark & 33.09 & 10.13 & & 25.35 & 7.84 \\
        Food and Entertaining & BMT & \cmark & 32.25 & 12.02 & & 21.86 & 7.66 \\
        Home and Garden & BMT & \cmark & 38.21 & 13.89 & & 29.04 & 11.86 \\
        Cars and Other Vehicles & BMT & \cmark & 33.09 & 14.79 & & 22.03 & 10.60 \\
        Holidays and Traditions & BMT & \cmark & 34.57 & 12.49 & & 26.72 & 10.00 \\
        Education and Communications & BMT & \cmark & 41.47 & 13.00 & & 23.95 & 7.07 \\
        Personal Care and Style & BMT & \cmark & 29.79 & 11.47 & & 22.25 & 7.32\\
        Pets and Animals & BMT & \cmark & 50.00 & 27.50 & & 42.04 & 16.86\\
        Health & BMT & \cmark & 42.52 & 22.30 & & 31.11 & 18.34 \\
        Family Life & BMT & \cmark & 35.71 & 14.29 & & 38.46 & 15.38 \\
        Arts and Entertainment & BMT & \cmark & 22.22 & 11.11 & & 11.76 & 5.88\\
        Sports and Fitness & BMT & \cmark & 33.62 & 20.13 & & 29.05 & 11.59\\

        \midrule
        Hobbies and Crafts & Joint (Ours) & \cmark & 38.11 & 13.63 & & 26.47 &    9.34 \\ 
        Food and Entertaining & Joint (Ours) & \cmark & 35.43 & 13.56 & & 28.54 & 10.4 \\ 
        Home and Garden & Joint (Ours) & \cmark & 35.28 & 17.63 & & 29.69 & 14.46 \\ 
        Cars and Other Vehicles & Joint (Ours) & \cmark & 39.68 & 15.23 & & 31.16 & 12.39 \\ 
        Holidays and Traditions & Joint (Ours) & \cmark & 34.52 & 10.92 & & 25.42 & 6.95 \\ 
        Education and Communications & Joint (Ours) & \cmark & 49.71 & 24.86 & & 31.61 & 13.94 \\
        Personal Care and Style & Joint (Ours) & \cmark & 41.38 & 17.57 & & 29.91 & 13.34 \\ 
        Pets and Animals & Joint (Ours) & \cmark & 70.0 & 30.0 & & 40.76 & 17.33 \\
        Health & Joint (Ours) & \cmark & 40.58 & 11.48 & & 37.26 & 7.42 \\
        Family Life & Joint (Ours) & \cmark & 50.0 & 28.57 & & 63.64 & 36.36 \\ 
        Arts and Entertainment & Joint (Ours) & \cmark & 22.22 & 11.11 & & 13.33 & 6.67 \\
        Sports and Fitness & Joint (Ours) & \cmark & 31.71 & 16.85 & & 24.84 & 9.5 \\ 
        \midrule
        All & BMT & \cmark & 34.06 & 12.34 & & 24.71 & 8.93 \\
        All & Joint (Ours) & \cmark & 37.50 & 14.76 & & 28.52 & 10.84 \\
        \bottomrule
    \end{tabular}
    }
    \caption{
        \Stepretrievalsubtask{} results per video category on our \dataname{} test split.
        \textit{FT: Finetuning on \dataname{}, Recall@IoU: Recall@1 with a threshold of IoU, Precision@IoU: Precision@1 with a threshold of IoU}.
    }
    \label{tab:moment_segmentation_categories}
\end{table}

\begin{table}[t]
    \centering
    \resizebox{\columnwidth}{!}{
    \begin{tabular}{l c c c c c c c}
        \toprule
        \multirow{2}{*}{\makecell{Moment \\ Duration}} & \multirow{2}{*}{Model} & \multirow{2}{*}{FT} & \multicolumn{2}{c}{Recall@IoU} & & \multicolumn{2}{c}{Precision@IoU}\\
        \cmidrule{4-5}
        \cmidrule{7-8}
        & & & 0.5 & 0.7 & & 0.5 & 0.7 \\       
        \midrule
        $<$ 1.5 mins & BMT & & 11.75 & 4.76 & & 26.92 & 9.19\\
        1.5 - 3 mins & BMT & & 6.59 & 3.31 & & 17.13 & 7.18\\
        $>$ 3 mins & BMT & & 6.40 & 3.04 & & 19.28 & 7.59 \\
        \midrule
        All & BMT & & 8.24 & 3.71 & & 20.95 & 7.96\\
        \midrule
        \midrule
        $<$ 1.5 mins & BMT & \cmark & 38.08 & 14.27 & & 27.75 & 10.87 \\
        1.5 - 3 mins & BMT & \cmark & 33.92 & 13.26 & & 23.20 & 8.74\\
        $>$ 3 mins & BMT & \cmark & 29.54 & 8.82 & & 23.23 & 6.90\\
        \midrule
        $<$ 1.5 mins & Joint (Ours) & \cmark & 44.32 & 17.81 & & 42.22 & 16.39 \\
        1.5 - 3 mins & Joint (Ours) & \cmark & 38.04 & 14.70 & & 25.41 & 9.68\\
        $>$ 3 mins & Joint (Ours) & \cmark & 28.72 & 11.27 & & 16.75 & 5.93 \\
        \midrule
        All & BMT & \cmark & 34.06 & 12.34 & & 24.71 & 8.93\\
        All & Joint (Ours) & \cmark & 37.50 & 14.76 & & 28.52 & 10.84\\
        \bottomrule
    \end{tabular}
    }
    \caption{
        \Stepretrievalsubtask{} results for different moment duration groups on our \dataname{} test split.
        \textit{FT: Finetuning on \dataname{}, Recall@IoU: Recall@1 with a threshold of IoU, Precision@IoU: Precision@1 with a threshold of IoU}.
    }
    \label{tab:moment_segmentation_durations}
\end{table}

\paragraph{\Stepretrievalsubtask{}.}
In \Cref{tab:moment_segmentation_categories}, we show the zeroshot and finetuned results of BMT~\cite{BMT_Iashin_2020} proposal module
and our joint model
on each individual category in our dataset. Finetuning BMT results show significant improvement in every category.

In \Cref{tab:moment_segmentation_durations}, we show the \stepretrievalsubtask{} performance on three moment duration groups.
All models achieve higher performance in shorter moments than in longer moments.
Our joint model shows better performance than BMT on shorter moments, while BMT does better on longer moments.

\begin{table*}[t]
    \centering
    \resizebox{\textwidth}{!}{
    \begin{tabular}{l c c c c c c c c}
        \toprule
        Category & Model & FT & METEOR & CIDEr & SPICE & Entailment (\%) & BERTScore & CLIPScore \\
        \midrule
        
        Hobbies and Crafts & SwinBERT & & 3.18 & 5.45 & 1.37 & 2.11 & 0.84 & 0.22 \\
        Food and Entertaining & SwinBERT & & 8.15 & 24.23 & 9.74 & 11.62 & 0.86 & 0.25\\
        Home and Garden & SwinBERT & & 3.23 & 6.35 & 1.34 & 1.49 & 0.84 & 0.21 \\
        Cars and Other Vehicles & SwinBERT & & 3.12 & 4.51 & 0.17 & 2.07 & 0.84 & 0.20  \\
        Holidays and Traditions & SwinBERT & & 4.84 & 10.55 & 3.26 & 2.28 & 0.84 & 0.22 \\
        Education and Communications & SwinBERT & & 3.16 & 7.96 & 2.58 & 3.41 & 0.83 & 0.24 \\
        Personal Care and Style & SwinBERT & & 4.48 & 16.05 & 4.83 & 4.69 & 0.84 & 0.22 \\
        Pets and Animals & SwinBERT & & 2.62 & 7.17 & 0 & 6.25 & 0.83 & 0.21  \\
        Health & SwinBERT & & 2.68 & 6.75 & 0.33 & 0& 0.83 & 0.19 \\
        Family Life & SwinBERT & & 2.46 & 9.78 & 0 & 14.29 & 0.83 & 0.20 \\
        Arts and Entertainment & SwinBERT & & 1.22 & 8.09 & 0 & 0 & 0.84 & 0.20 \\
        Sports and Fitness & SwinBERT & & 1.87 & 3.59 & 0 & 6.82 & 0.84 & 0.23 \\
        \midrule
        All & SwinBERT & & 5.12 & 13.31 & 4.65 & 5.86 & 0.85 & 0.23 \\
        \midrule
        \midrule
        Hobbies and Crafts & SwinBERT & \cmark & 4.54 & 13.82 & 4.88 & 38.95 & 0.86 & 0.23 \\
        Food and Entertaining & SwinBERT & \cmark & 8.08 & 37.64 & 9.82 & 32.60 & 0.87 & 0.24  \\
        Home and Garden & SwinBERT & \cmark & 4.84 & 18.04 & 5.26 & 41.04 & 0.86 & 0.22 \\
        Cars and Other Vehicles & SwinBERT & \cmark & 5.49 & 21.59 & 6.64 & 27.80 & 0.87 & 0.22 \\
        Holidays and Traditions & SwinBERT & \cmark & 5.52 & 20.56 & 4.29 & 32.42 & 0.86 & 0.23 \\
        Education and Communications & SwinBERT & \cmark & 3.45 & 10.51 & 1.33 & 23.86 & 0.85 & 0.24 \\
        Personal Care and Style & SwinBERT & \cmark & 5.86 & 25.79 & 6.25 & 39.84 & 0.86 & 0.22 \\
        Pets and Animals & SwinBERT & \cmark & 4.59 & 18.74 & 0 & 21.25 & 0.86 & 0.22 \\
        Health & SwinBERT & \cmark & 2.27 & 5.06 & 0 & 10.91& 0.85 & 0.20 \\
        Family Life & SwinBERT & \cmark & 4.96 & 9.52 & 3.57 & 42.86 & 0.85 & 0.22 \\
        Arts and Entertainment & SwinBERT & \cmark & 2.90 & 13.16 & 0 & 11.11 & 0.85 & 0.21 \\
        Sports and Fitness & SwinBERT & \cmark & 2.65 & 12.78 & 0.91 & 54.55 & 0.86 & 0.24 \\
        \midrule
        
        Hobbies and Crafts & Joint (Ours) & \cmark & 3.98 & 18.26 & 3.61 & 35.31  & 0.86 & 0.23 \\ 
        Food and Entertaining & Joint (Ours) & \cmark & 4.22 & 30.35 & 2.63 & 57.67 & 0.86 & 0.23 \\
        Home and Garden & Joint (Ours) & \cmark & 4.24 & 14.88 & 4.43 & 34.33 & 0.86 & 0.22 \\ 
        Cars and Other Vehicles & Joint (Ours) & \cmark & 5.41 & 22.20 & 6.65 & 28.33 & 0.87 & 0.23 \\ 
        Holidays and Traditions & Joint (Ours) & \cmark & 4.40 & 22.83 & 3.48 & 38.81 & 0.85 & 0.22 \\
        Education and Communications & Joint (Ours) & \cmark & 4.01 & 19.37 & 4.17 & 27.27 & 0.85 & 0.23\\
        Personal Care and Style & Joint (Ours) & \cmark & 3.37 & 18.09 & 4.74 & 57.03 & 0.85 & 0.23 \\ 
        Pets and Animals & Joint (Ours) & \cmark & 3.55 & 13.99 & 3.12 & 12.50 & 0.85 & 0.23 \\ 
        Health & Joint (Ours) & \cmark & 2.33 & 11.37 & 2.42 & 30.91 & 0.85 & 0.19 \\ 
        Family Life & Joint (Ours) & \cmark & 3.68 & 3.41 & 7.14 & 35.71 & 0.84 & 0.22 \\ 
        Arts and Entertainment & Joint (Ours) & \cmark & 1.67 & 0 & 10.00 & 11.11  & 0.84 & 0.21 \\ 
        Sports and Fitness & Joint (Ours) & \cmark & 2.54 & 17.79 & 2.65 & 40.91 & 0.86 & 0.23 \\ 
        \midrule
        All & SwinBERT & \cmark & 5.94 & 24.66 & 6.67 & 35.09 & 0.86 & 0.23 \\
        All & Joint (Ours) & \cmark & 4.13 & 23.01 & 3.54 & 43.88 & 0.86 & 0.23 \\

        \bottomrule
    \end{tabular}
    }
    \caption{
        \Momentcaptioningtask{} results per video category on our \dataname{} test split.
        \textit{FT: Finetuning on \dataname{}}.
    }
    \label{tab:step_captioning_categories}
\end{table*}

\begin{table*}[t]
    \centering
    \resizebox{\textwidth}{!}{
    \begin{tabular}{l c c c c c c c c c c c }
        \toprule
        Step Duration & Model & FT & METEOR & CIDEr & SPICE & Entailment (\%) & BERTScore & CLIPScore \\
        \midrule
        $<$ 8 secs & SwinBERT & & 5.73 & 16.72 & 5.40 & 6.74 & 0.84 & 0.23 \\
        8 - 18 secs & SwinBERT & & 5.18 & 13.95 & 4.95 & 6.05 & 0.85 & 0.23 \\
        $>$ 18 secs & SwinBERT & & 4.57 & 10.66 & 3.66 & 4.96 & 0.84 & 0.23\\
        \midrule
        All & SwinBERT & & 5.12 & 13.31 & 4.65 & 5.86 & 0.85 & 0.23 \\
        \midrule
        \midrule
        $<$ 8 secs & SwinBERT & \cmark & 6.25 & 25.32 & 6.94 & 25.37 & 0.86 & 0.23 \\
        8 - 18 secs & SwinBERT & \cmark & 6.21 & 25.92 & 6.31 & 32.99 & 0.86 & 0.23\\
        $>$ 18 secs & SwinBERT & \cmark & 5.40 & 23.64 & 6.83 & 37.03 & 0.86 & 0.23\\
        \midrule
        $<$ 8 secs & Joint (Ours) & \cmark & 4.22 & 22.49 & 3.24 & 48.67 & 0.85 & 0.22 \\
        8 - 18 secs & Joint (Ours) & \cmark & 4.02 & 22.55 & 3.36 & 41.25 & 0.86 & 0.23 \\
        $>$ 18 secs & Joint (Ours) & \cmark & 4.17 & 24.83 & 4.02 & 41.29 & 0.86 & 0.22\\
        
        \midrule
        All & SwinBERT & \cmark & 5.94 & 24.66 & 6.67 & 35.09 & 0.86 & 0.23 \\
        All & Joint (Ours) & \cmark & 4.13 & 23.01 & 3.54 & 43.88 & 0.86 & 0.23 \\
        \bottomrule
    \end{tabular}
    }
    \caption{
        \Momentcaptioningtask{} results for various step durations on our \dataname{} test split.
        \textit{FT: Finetuning on \dataname{}}.
    }
    \label{tab:step_captioning_durations}
\end{table*}

\paragraph{\Momentcaptioningtask{}.}
In \Cref{tab:step_captioning_categories}, we show the zeroshot and finetuned results of SwinBERT~\cite{lin2021SwinBERTend-to-end}
and our joint model
on each video category in our dataset.
Notably, SwinBERT performs best in the `Food and Entertaining' category, likely because SwinBERT was pretrained on the YouCook2~\cite{Zhou2018TowardsALYouCook2} dataset.
\par
In \Cref{tab:step_captioning_durations}, we show the \momentcaptioningtask{} performance in different step durations.
Both N-gram (\eg CIDEr) and sentence-level embedding metrics (BERTScore and CLIPScore) do not show significant differences among different categories.
In the entailment metric,
finetuned SwinBERT gets better as the steps get longer,
while our joint model gets slightly worse for longer steps.

\section{Evaluation Details}
\label{sec:evaluation_detail}

We continue the evaluation details of \stepretrievalsubtask{} task in the main paper \cref{sec:metrics} (Metrics).
The BMT~\cite{BMT_Iashin_2020} model generates up to 100 possible step segments, where many of them are outside of the ground-truth (GT) moment input, overlap each other, and there are also gaps between segments.
For evaluation of \stepretrievalsubtask{} task, we first remove any segments outside the given ground truth moment and use non-maximum suppression (NMS) to remove any overlapping segments.
Then any resulting gaps between steps are also marked as separate steps.

\end{document}